\documentclass{article} 
\usepackage{collas2022_conference,times}

\usepackage{amsmath,amsfonts,bm}

\def\eqref#1{equation~\ref{#1}}

\def\1{\bm{1}}

\def\rvtheta{{\mathbf{\theta}}}

\def\rvx{{\mathbf{x}}}
\def\rvy{{\mathbf{y}}}

\DeclareMathAlphabet{\mathsfit}{\encodingdefault}{\sfdefault}{m}{sl}
\SetMathAlphabet{\mathsfit}{bold}{\encodingdefault}{\sfdefault}{bx}{n}

\def\sN{{\mathbb{N}}}

\def\sR{{\mathbb{R}}}

\newcommand{\E}{\mathbb{E}}

\DeclareMathOperator*{\argmax}{arg\,max}
\DeclareMathOperator*{\argmin}{arg\,min}

\usepackage{hyperref}
\hypersetup{
    colorlinks=true,
    linkcolor=red,
    filecolor=magenta,      
    urlcolor=blue,
    citecolor=purple,
    pdftitle={Overleaf Example},
    pdfpagemode=FullScreen,
    }

\title{Energy-Based Models for Continual Learning}

\author{Shuang Li \thanks{Correspondence to: Shuang Li $<$lishuang@mit.edu$>$} \\
MIT CSAIL \\
\texttt{lishuang@mit.edu} \\
\And 
Yilun Du  \\
MIT CSAIL \\
\texttt{yilundu@mit.edu} \\
\And
Gido M. van de Ven \\
Baylor College of Medicine \\
\texttt{ven@bcm.edu} \\
\And 
Igor Mordatch \\
Google Brain \\
\texttt{igor.mordatch@gmail.com}
}

\collasfinalcopy 

\usepackage{booktabs}
\usepackage{colortbl}

\usepackage{float,wrapfig,blindtext}

\usepackage{array}
\usepackage{adjustbox}
\usepackage{multirow}
\usepackage{colortbl}
\usepackage{caption}
\usepackage{amsmath}
\usepackage{subcaption}
\usepackage{multicol}
\usepackage{mathtools}
\usepackage{hyperref}

\makeatletter
\DeclareRobustCommand\onedot{\futurelet\@let@token\@onedot}
\def\@onedot{\ifx\@let@token.\else.\null\fi\xspace}

\def\eg{\emph{e.g}\onedot} 
\def\ie{\emph{i.e}\onedot}

\makeatother

\newcommand{\sect}[1]{Section~\ref{#1}}

\newcommand{\eqn}[1]{Equation~(\ref{#1})}
\newcommand{\fig}[1]{Figure~\ref{#1}}
\newcommand{\tbl}[1]{Table~\ref{#1}}

\newcommand{\ignore}[1]{}

\definecolor{rowblue}{RGB}{220,230,240}
\definecolor{myorchid}{RGB}{150,10,30}
\definecolor{myblue}{RGB}{10,30,250}
\definecolor{mygreen}{RGB}{10,120,10}

\begin{document}

\maketitle

\begin{abstract}
We motivate Energy-Based Models (EBMs) as a promising model class for continual learning problems. Instead of tackling continual learning via the use of external memory, growing models, or regularization, EBMs change the underlying training objective to cause less interference with previously learned information. Our proposed version of EBMs for continual learning is simple, efficient, and outperforms baseline methods by a large margin on several benchmarks. Moreover, our proposed contrastive divergence-based training objective can be combined with other continual learning methods, resulting in substantial boosts in their performance. We further show that EBMs are adaptable to a more general continual learning setting where the data distribution changes without the notion of explicitly delineated tasks. These observations point towards EBMs as a useful building block for future continual learning methods. Project page: \href{https://energy-based-model.github.io/Energy-Based-Models-for-Continual-Learning/}{\small{https://energy-based-model.github.io/Energy-Based-Models-for-Continual-Learning/}}.
\end{abstract}

\section{Introduction}
Humans are able to rapidly learn new skills and continuously integrate them with prior knowledge. The field of Continual Learning (CL) seeks to build artificial agents with the same capabilities \citep{parisi2019continual,hadsell2020embracing,delange2021continual}.
In recent years, continual learning has seen increased attention, particularly in the context of classification problems.
Continual learning requires models to remember prior skills as well as incrementally learn new skills, without necessarily having a notion of an explicit task identity.
Standard neural networks \citep{he2016deep,simonyan2014very,szegedy2015going} experience  catastrophic forgetting and perform poorly in this setting.
Different approaches have been proposed to mitigate catastrophic forgetting, but many rely on the usage of external memory \citep{lopez2017gradient,li2017learning,hayes2020remind,buzzega2020dark}, additional models \citep{shin2017continual,von2019continualP,liu2020generative}, or auxiliary objectives and regularization \citep{kirkpatrick2017overcoming,schwarz2018progress,maltoni2019continuous,zenke2017continual}, which can constrain the wide applicability of these methods. 

In this work, we propose a new approach towards continual learning on classification tasks.
Most existing CL approaches tackle these tasks 
by utilizing normalized probability distribution (i.e., softmax output layer) and trained with a cross-entropy objective. In this paper, we argue that by viewing classification from the lens of training an un-normalized probability distribution, we can significantly improve continual learning performance in classification problems.  In particular, we interpret classification as learning an Energy-Based Model (EBM) across  classes.
Training becomes a wake-sleep process, where the energy of an input data at its ground truth label is decreased while the energy of the input at (an)other selected class(es) is increased. An important advantage is that this framework offers freedom to choose what classes to update in the continual learning process.
By contrast, the cross entropy objective reduces the likelihood of \emph{all} negative classes when given a new input, creating updates that lead to catastrophic forgetting. 

The energy function, which maps an input-label pair to a scalar energy, also provides a way for the model to select and filter portions of the input that are relevant towards the classification on hand. This enables EBMs training updates for new data to interfere less with previous data. In particular, our formulation of the energy function allows us to compute the energy of an input by learning a conditional gain based on the class label, which serves as an attention filter to select the most relevant information. In the event of a new class, a new conditional gain  can be learned. 

These unique properties benefit EBMs in addressing two important open challenges in continual learning.
1)~First, we show that EBMs are promising for class-incremental learning, which is one of the most challenging settings for continual learning \citep{van2019three,hayes2020lifelong,masana2020class,prabhu12356gdumb}. Generally, successful existing methods for class-incremental learning store data, use generative replay, or pretrain their model on another dataset, which has disadvantages in terms of memory and/or computational efficiency. We show that EBMs perform well in class-incremental learning without using replay and without relying on stored data.
2) The second open challenge that EBMs can address is continual learning without task boundaries~\citep{aljundi2019online,lee2019neural,jin2020gradient}. Typically, a continual learning problem is set up as a sequence of distinct tasks with clear boundaries that are known to the model (the \emph{boundary-aware} setting). Most existing continual learning methods rely on these known boundaries for performing certain consolidation steps (e.g., calculating parameter importance, updating a stored copy of the model). However, assuming such clear boundaries is not always realistic, and often a more natural scenario is the \emph{boundary-free} setting \citep{zeno2018task,rajasegaran2020itaml}, in which data distributions gradually change without a clear notion of task boundaries. 
While many common CL methods cannot be used without clear task boundaries, we show that EBMs can be naturally applied to this more challenging setting.

There are three main contributions of our work. 
First, we introduce energy-based models for classification continual learning problems. We show that EBMs can naturally deal with challenging problems in CL, including the boundary-free setting and class-incremental learning without using replay.
Secondly, we propose an energy-based training objective that is \textit{simple} and broadly applicable to different types of models, with significant boosts on their performance. 
This contrastive divergence based training objective can naturally handle the dynamically growing number of classes and significantly reduces catastrophic forgetting. 
Lastly, we show that the proposed EBMs perform strongly on four standard CL benchmarks: split MNIST, permuted MNIST, CIFAR-10, and CIFAR-100 without using extra generative model or storing data.
These observations point towards EBMs as a class of models naturally inclined towards the CL regime and as an important new baseline upon which to build further developments.


\section{Related Work}
\label{sec:related_work}
\subsection{Continual learning settings}
\label{sec:continual_settings}
\noindent \textbf{Boundary-aware versus boundary-free during training}.
In most existing CL studies, models are trained in a \textit{boundary-aware} setting, in which a sequence of distinct tasks with clear task boundaries is given (\eg, \citep{kirkpatrick2017overcoming,zenke2017continual,shin2017continual}). 
There are no overlaps between any two tasks.
Models are first trained on the first task and then move to the second one. 
Moreover, models are typically told when there is a transition from one task to the next. 
However, it could be argued that it is more realistic for tasks to change gradually and for models to not be explicitly informed about the task boundaries. 
Such a \textit{boundary-free} setting has been explored in \citep{zeno2018task,rajasegaran2020itaml,aljundi2019task,madireddy2020neuromodulated}. In this setting, models learn in a streaming fashion and the data distributions gradually change over time. 
Importantly, most existing CL methods are not applicable to this setting as they require the task boundaries to decide when to perform certain consolidation steps.
In this paper, we show that our method can naturally handle both the boundary-aware and boundary-free settings.

\noindent \textbf{Task-incremental versus class-incremental learning}.
Another important distinction in CL is between task-incremental learning (\textit{Task-IL}) and class-incremental learning (\textit{Class-IL}) \citep{van2019three,prabhu12356gdumb,belouadah2020initial,maltoni2019continuous,hayes2020lifelong}. 
In \textit{Task-IL}, also referred to as the multi-head setting \citep{farquhar2018towards}, models predict the label of an input data by choosing only from the labels in the task where the data come from. 
In \textit{Class-IL}, also referred to as the single-head setting, models chose between the classes from all tasks so far when asked to predict the label of an input data.
\textit{Class-IL} is more challenging than \textit{Task-IL} as it requires models to select the correct labels from the mixture of new and old classes. 

In some previous works, class-incremental learning only refers to the boundary-aware setting where the tasks have clear boundaries. In our paper, we use class-incremental learning in a more general way, allowing it to contain both boundary-aware and boundary-free settings, where the tasks do not necessarily have clear boundaries.


\vspace{-5pt}
\subsection{Continual learning approaches} 

\noindent \textbf{Task-specific methods.}
One way to reduce interference between tasks is by using different parts of a neural network for different tasks~\citep{fernando2017pathnet,serra2018overcoming,masse2018alleviating,zeng2019continual,hu2019overcoming,wortsman2020supermasks}.
Other methods let models grow or recruit new resources when learning new tasks~\citep{rusu2016progressive,yoon2017lifelong,vogelstein2020omnidirectional}.
Although these methods are generally successful in reducing catastrophic forgetting, a key disadvantage is that they require knowledge of task identities during training and testing. Unless they are combined with a mechanism for task-inference, they are not suitable for \textit{Class-IL}.

\noindent \textbf{Regularization-based methods.}
Regularization is used in CL to encourage the stability of those aspects of the network that are important for previous tasks \citep{li2017learning,zenke2017continual,nguyen2017variational,titsias2020functional,kolouri2020sliced,kao2021natural}. 
A popular strategy is to add a regularization loss to penalise changes of important parameters.
For example, EWC \citep{kirkpatrick2017overcoming} and online EWC \citep{schwarz2018progress} evaluate the importance of parameters based on the fisher information matrices.
A disadvantage of regularization-based methods is that typically they gradually reduce the model's capacity for learning new tasks. 
Moreover, while in theory these methods can be used for \textit{Class-IL}, in practice they have been shown to fail on such problems \citep{farquhar2018towards,van2019three}.

\noindent \textbf{Replay methods.}
To preserve knowledge, replay methods periodically rehearse previous information during training \citep{robins1995catastrophic,rebuffi2017icarl,aljundi2019online,chaudhry2019tiny}. 
Exact or experience replay based methods store data from previous tasks and revisit them when training on new tasks. Although straightforward, such methods face critical non-trivial questions, such as how to select the data to be stored and how to use them \citep{lopez2017gradient,hou2019learning,wu2019large,chaudhry2019efficient,mundt2020wholistic,pan2020continual}.
An alternative is to generate the replayed data~\citep{shin2017continual}. 
While both types of replay can mitigate forgetting, an important disadvantage is that they are computationally relatively expensive.
Additionally, storing data might not always be possible while incrementally training a generative model is a challenging problem in itself \citep{lesort2019generative,vandeven2020brain}.

In contrast, we propose EBMs for CL that reduce catastrophic forgetting without requiring knowledge of task-identity, without restricting the model’s learning capabilities, and without using stored data.

\section{Rethinking Existing Continual Learning}
\label{sec:background}
The most common way to do classification with deep neural networks is to use a softmax output layer in combination with a cross-entropy loss.
In continual learning, most existing methods for classification are based on the softmax-based classifier (SBC). 

\subsection{Softmax-based classification}
\label{sec:softmax}
Given an input $\rvx \in \sR^D$ and a discrete set $\mathcal{Y}=\{1, \dots, N\}$ of $N$ possible class labels, a traditional softmax-based classifier defines the conditional probabilities of those labels as:
\begin{equation}
\small
    p_{\boldsymbol{\theta}}(y | \rvx) = \frac{\text{exp}({[f_{\boldsymbol{\theta}}(\rvx)]}_{y})} {\sum_{i\in \mathcal{Y}}{\text{exp}({[f_{\boldsymbol{\theta}}(\rvx)]}_{i})}},
    \;\;\;
    \text{for all } y \in \mathcal{Y},
\label{eqn:softmax}
\end{equation}
where $f_{\boldsymbol{\theta}}(\rvx): \sR^D \rightarrow \sR^N$ is a feed-forward neural network, parameterized by $\boldsymbol{\theta}$, that maps an input $\rvx$ to a $N$-dimensional vector of logits. 
$[\cdot]_i$ indicates the $i^{\text{th}}$ element of a vector. 

\noindent\textbf{Training.} A softmax-based classifier is typically trained by optimizing the cross-entropy loss function. For a given input $\rvx$ and corresponding ground truth label $y^+$, the cross-entropy loss is $\mathcal{L}_{CE}(\boldsymbol{\theta};\rvx,y^+) = - \text{log}(p_{\boldsymbol{\theta}}(y^+|\rvx))$. 
A schema of SBC is shown in the left part of \fig{fig:ebm_architecture_paper}.

\noindent\textbf{Inference.} Given an input $\rvx$, the class label predicted by the softmax-based classifier is the class with the largest conditional probability $\hat{y} = \argmax_{y\in \mathcal{Y}} p_{\boldsymbol{\theta}}(y|\rvx)$.


\subsection{Why are softmax-based classifiers not the best way to do continual learning?}
\label{sec:feed_forward_classifier}
When used for continual learning, and in particular when used for class-incremental learning, softmax-based classifiers face several challenges. While the cross-entropy loss is designed for i.i.d. datapoints and labels, the data found in continual learning does not follow such a distribution.  As a result, when training on a new task, the likelihood of the currently observed classes is increased, but the likelihood of old classes is too heavily suppressed since they are not encountered in the new task. The softmax operation itself introduces competitive, winner-take-all dynamics that make the classifier catastrophically forget past tasks.
We show such phenomenon of SBC in \sect{sec:qualitative}.


\section{Continual learning with Energy-Based Models}
In this section, we propose a simple but efficient energy-based training objective that can successfully mitigate the catastrophically forgetting in CL.
We first introduce EBMs in section~\ref{sec:ebms} and then show how EBMs are used for classification in section~\ref{sec:method_ebm_classification} and continual learning in sections~\ref{sec:method_ebm_classification_cl}.


\subsection{Energy-based models}
\label{sec:ebms}
EBMs \citep{lecun2006tutorial} are a class of maximum likelihood models that define the likelihood of a data point $\rvx \in \mathcal{X} \subseteq \sR^D$ using the Boltzmann distribution with the partition function $Z(\boldsymbol{\theta})$:
\vspace{-3pt}
\begin{equation}
\small
    p_{\boldsymbol{\theta}}(\rvx) = \frac{\exp(-E_{\boldsymbol{\theta}}(\rvx))}{Z(\rvtheta)},
    \;\;\;\;
    Z(\boldsymbol{\theta}) = \int_{\rvx\in\mathcal{X}} \text{exp}(-E_{\boldsymbol{\theta}}(\rvx))
\end{equation}
where $E_{\boldsymbol{\theta}}(\rvx): \sR^D \rightarrow \sR$, known as the energy function, maps each data point $\rvx$ to a scalar energy value. In deep learning applications, the energy function $E_{\boldsymbol{\theta}}$ is a neural network parameterized by $\boldsymbol{\theta}$.

EBMs are powerful models that have been applied to different domains, such as structured prediction \citep{belanger2016structured,rooshenas2019search,tu2019benchmarking}, text generation \citep{deng2020residual}, RL \citep{haarnoja2017reinforcement}, image generation \citep{pmlr-v5-salakhutdinov09a,du2019implicit,du2020compositional,nijkamp2019anatomy}, and classification \citep{grathwohl2019your}. An important difference from prior EBM methods for image classification \citep{grathwohl2019your} is that we do not utilize negative samples from MCMC in the training objective.
Utilizing MCMC to sample negative images makes EBMs difficult to scale to complex datasets and makes training much slower. Our method that uses negative labels in the training objective is faster and more scalable.
In this paper, we study continual learning on classification tasks. As far as we are aware, EBMs for continual learning has so far remained unexplored. 

\begin{wrapfigure}{r}{0.5\textwidth}
\vspace{-20pt}
  \begin{center}
    \includegraphics[width=1\linewidth]{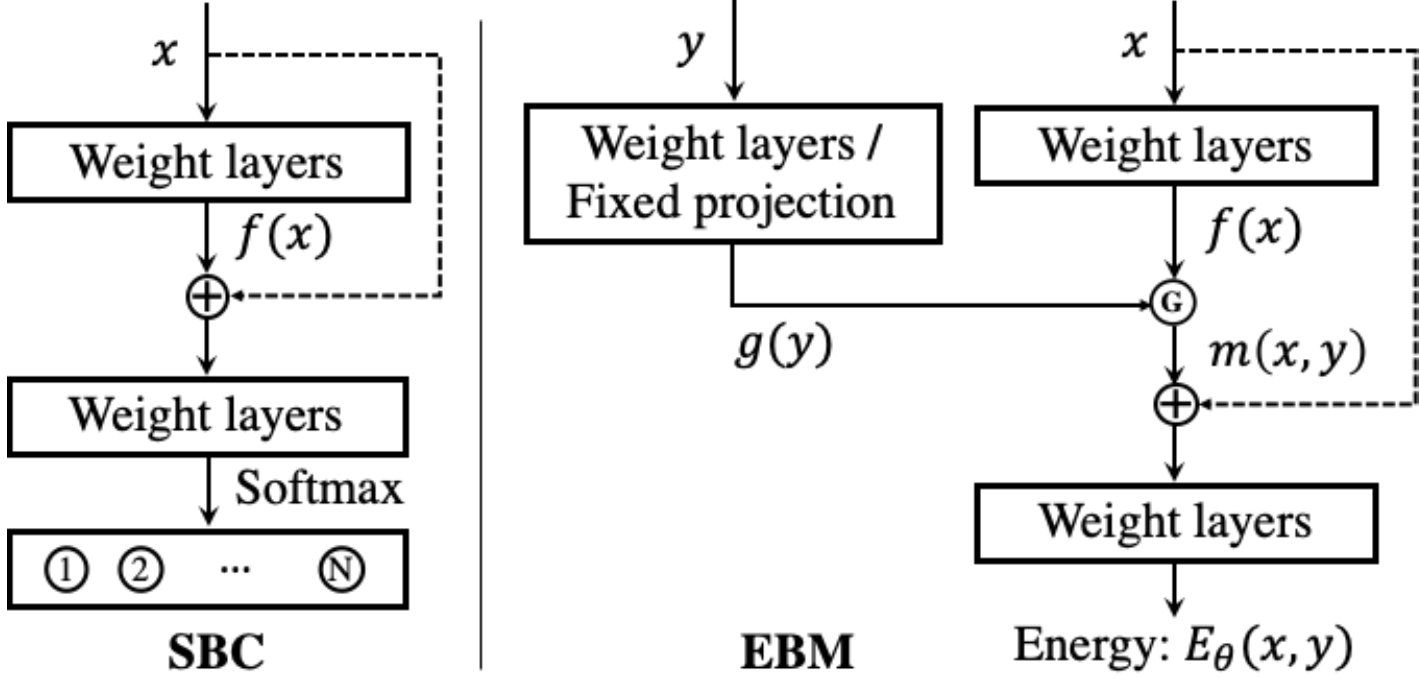}
  \end{center}
  \vspace{-10pt}
  \caption{\small{Schematic of the model architectures of the softmax-based classifier (SBC) and energy-based models (EBM). SBC takes an image $\rvx$ as input and outputs a fixed pre-defined $N$-dimensional vector. 
    EBM takes a data $\rvx$ and a class $y$ as input and outputs their energy value. The dash lines are optional skip connections.}}
    \label{fig:ebm_architecture_paper}
\vspace{-10pt}
\end{wrapfigure}

\subsection{Energy-based models for classification}
\label{sec:method_ebm_classification}


To solve the classification tasks, we adapt the above general formulation of an EBM as follows.
Given inputs $\rvx\in \sR^D$ and a discrete set $\mathcal{Y}$ of possible class labels, 
%
we propose to use the Boltzmann distribution to define the conditional likelihood of label $y$ given $\rvx$:
\begin{equation}
\small
    \label{eqn:likelihood}
    p_{\boldsymbol{\theta}}(y | \rvx) 
    = \frac{\text{exp}(- E_{\boldsymbol{\theta}}(\rvx, y))} {Z(\boldsymbol{\theta}; \rvx)}, 
    Z(\boldsymbol{\theta}; \rvx)
    =\sum_{y' \in\mathcal{Y}} \text{exp}(- E_{\boldsymbol{\theta}}(\rvx, y'))
\end{equation}
where $E_{\boldsymbol{\theta}}(\rvx, y): (\sR^D,\sN) \rightarrow \sR$ is the energy function that maps an input-label pair $(\rvx,y)$ to a scalar energy value, and $Z(\boldsymbol{\theta}; \rvx)$ is the partition function for normalization.

\noindent\textbf{Training.} 
We want the distribution defined by $E_{\boldsymbol{\theta}}$ to model the data distribution $p_D$, which we do by minimizing the negative log likelihood of the data $\mathcal{L}_{\text{ML}}(\boldsymbol{\theta}) = \E_{(x,y) \sim p_D} [ - \text{log} p_{\boldsymbol{\theta}} (y | \rvx) ]$ with the expanded form:
\vspace{-3pt}
\begin{equation}
\small
    \mathcal{L}_{\text{ML}}(\boldsymbol{\theta}) 
    = 
    \E_{(\rvx, y) \sim p_D} 
    \left [
    E_{\boldsymbol{\theta}}(\rvx, y) + \log (\sum_{y' \in \mathcal{Y} } e^{-E_{\boldsymbol{\theta}}(\rvx, y')}) 
    \right].
\label{eqn:ml_loss_expend}
\end{equation}
\eqn{eqn:ml_loss_expend} minimizes the energy of $\rvx$ at the ground truth label $y$ and minimizes the overall partition function by increasing the energy of $\rvx$ at other labels $y'$.

\noindent \textbf{Inference.}
Given an input $\rvx$, the class label predicted by our EBMs is the class with the smallest energy at $\rvx$, which can be written as $\hat{y} = \argmin_{y' \in\mathcal{Y}} E_{\rvtheta}(\rvx, y')$.

\subsection{Energy-based model as a building block for continual learning}
\label{sec:method_ebm_classification_cl}
\noindent\textbf{EBM training objective.}
\label{sec:ebm_training_objective}
In \eqn{eqn:ml_loss_expend}, the energy across all class labels $y'$ given data $\rvx$ is maximized. 
Directly maximizing energy across all labels raises the same problem as the softmax-based classifier models that the old classes are suppressed when training a model on new classes and thus cause the catastrophic forgetting. 
Inspired by \citep{hinton2002training}, we find that the contrastive divergence approximation of \eqn{eqn:ml_loss_expend} can mitigate this problem and lead to a simpler equation.
%
To do so, we define the following contrastive divergence loss:
\begin{equation}
\small
    \label{eqn:cd_loss}
    \mathcal{L}_{\text{CD}}(\boldsymbol{\theta}; \rvx, y) = \E_{(\rvx,y) \sim p_D} 
    \left[ E_{\boldsymbol{\theta}}(\rvx,y) - E_{\boldsymbol{\theta}}(\rvx, y^-) \right],
\end{equation}
where $y$ is the ground truth label of data  $\rvx$ and $y^-$ is a negative class label randomly sampled from the set of class labels in the current training batch $\mathcal{Y}_B$ such that $y^- \neq y$.

Different from the softmax-based classifier, EBMs maximize likelihood not by normalizing over all classes but instead by contrastively increasing the energy difference between the ground truth label and another negative label for a given data point. This operation causes less interference with previous classes and enables EBMs to suffer less from catastrophic forgetting.

Importantly, such a sampling strategy also allows EBMs to be naturally applied, without any modification, to different CL settings (\sect{sec:continual_settings}). 
Since we select the negative sample(s) from the current batch, our EBMs do not require knowledge of the task on hand. This allows application of EBMs in the \emph{boundary-free} setting, in which the underlying tasks or task boundaries are not given. See Appendix~\ref{apx:boundary-free setting} for more details about the \emph{boundary-free} setting.

We find that the proposed EBM training objective is efficient enough to achieve good performance on different CL datasets (\tbl{tab:comp_baselines_boundary_aware} and \tbl{tab:comp_baselines_boundary_free}).
We note however that it is possible to use other strategies for choosing the negative classes in the partition function in \eqn{eqn:ml_loss_expend}.
In \tbl{tab:neg_sam}, we explore alternative strategies: 1) using all classes in the current training batch $\mathcal{Y_B}$ as negative classes, and
2) using all classes seen so far as negative classes.
The usage of negative samples in the EBM training objective provides freedom for models to choose which classes to train on 
which is important for preventing catastrophic forgetting in CL.

\noindent\textbf{Energy network.}
\label{sec:network_structure}
Another important difference from softmax-based classifiers is that the choice of model architectures becomes more flexible in EBMs. Traditional classification models only feed in $\rvx$ as input. 
In contrast, EBMs have many different ways to combine $\rvx$ and $y$ in the energy function with the only requirement that $E_{\theta}(\rvx, y): (\sR^D,\sN) \rightarrow \sR$.
In EBMs, we can treat $y$ as an attention filter or gate to select the most relevant information between $\rvx$ and $y$.

To compute the energy of any data $\rvx$ and class label $y$ pair, we use $y$ to influence a conditional gain on $\rvx$, which serves as an attention filter \citep{xu2015show} to select the most relevant information between $\rvx$ and $y$.
In \fig{fig:ebm_architecture_paper} (right), we first send $\rvx$ into a network to generate a feature $f(\rvx)$. The label $y$ is mapped into a same dimension feature $g(y)$ using a small learned network or a fixed projection. 
We use the gating block $G$ to select the most relevant information between $\rvx$ and $y$:
\begin{equation}
m(\rvx,y) = G(f(\rvx), g(y)).
\label{eqn:gain}
\end{equation}
The output is finally sent to weight layers to generate the energy value $E_{\theta}(\rvx,y)$. 
See Appendix \ref{appendix:model_architectures} for more details about our model architectures.

Our EBMs allow any number of classes in new batches by simply training or defining a new conditional gain $g(\rvy)$ for the new classes and generating its energy value with data point $\rvx$. 


\noindent\textbf{Inference.}
\label{sec:interence_ebm}
During inference, because we evaluate according to the class-incremental learning scenario, the model must predict a class label by choosing from all classes seen so far.
Let $\rvx$ be one data point with an associated discrete label $y$.
There are $\mathcal{Y}$ different class labels in the dataset. 
The MAP estimate is 
$\hat{y} = \arg\min_{y'} E_{\theta}(\rvx, y')$, 
where $y' \in \mathcal{Y}$ and $E_{\theta}$ is the energy function with parameters $\theta$ after training on all the training data.

\noindent\textbf{Alternative EBM training objective.}
\label{sec:alternative_training_objective}
EBMs are not limited to modeling the conditional distribution between $\rvx$ and $y$ as shown in \eqn{eqn:likelihood}. Another way to use the Boltzmann distribution is to define the joint likelihood of $\rvx$ and $y$. 
In Appendix \ref{apx:alternative_training_objective_appendix}, we show that modeling the joint likelihood can further improve the results.
Since the main focus of this paper is to propose a simple but efficient EBM training objective for continual learning, we only show the results of using \eqn{eqn:cd_loss} in this main paper.

\section{Experiments}
\label{sec:experiments}

In this section, we want to answer the following questions: How do the proposed EBMs perform on different CL settings? Can we apply the EBM training objective to other methods? And what is the best architecture for label conditioning? 
To answer these questions, we first report experiments on the \emph{boundary-aware} setting in \sect{exp:boundary-aware}. We then show that EBMs can also be applied to the \emph{boundary-free} setting in \sect{exp:boundary-free}.

\subsection{Experiments on Boundary-Aware Setting}
\label{exp:boundary-aware}


\noindent\textbf{Datasets.}
We evaluate the proposed EBMs on four commonly used CL datasets, including split MNIST \citep{zenke2017continual}, permuted MNIST \citep{kirkpatrick2017overcoming}, CIFAR-10 \citep{krizhevsky2009learning}, and CIFAR-100 \citep{krizhevsky2009learning}. 
The split MNIST is obtained by splitting the original MNIST \citep{lecun1998mnist} into 5 tasks with each task having 2 classes. It has 60,000 training images and 10,000 test images. 
We follow the implementation of \citep{van2019three} for permuted MNIST. There are 10 tasks and each task has 10 classes.
We separate CIFAR-10 into 5 tasks, each task with 2 classes. 
CIFAR-100 is split into 10 tasks with each task having 10 classes. 
CIFAR-10 and CIFAR-100 each has 50,000 training images and 10,000 test images.

\begin{table}
    \caption{\small Evaluation of class-incremental learning on the \emph{boundary-aware} setting on four datasets. 
    Experiments ran by ourselves are performed 10 times with different random seeds, the results are reported as the mean $\pm$ SEM over these runs. 
    Results with (-) means they are not reported in their original paper.
    Results with $\pm$ (-) means a SEM is not reported in their original paper.
    }
    \label{tab:comp_baselines_boundary_aware}
    \vspace{-8pt}
    \begin{center}
    \scalebox{0.82}{
    \begin{tabular}{lcccc}
    \toprule
    \bf Without Replay & \bf split MNIST & \bf permuted MNIST & \bf CIFAR-10 & \bf CIFAR-100 \\
    \midrule
    SBC & 19.90 $\pm$ 0.02 & 17.26 $\pm$ 0.19 & 19.06 $\pm$ 0.05 & 8.18 $\pm$ 0.10 \\
    \bf EBM & \bf 53.12 $\pm$ \bf 0.04 & \bf 87.58 $\pm$ 0.50 & \bf 38.84 $\pm$ 1.08 & \bf 30.28 $\pm$ 0.28  \\
    \midrule
    EWC \citep{kirkpatrick2017overcoming} & 20.01 $\pm$ 0.06 & 25.04 $\pm$ 0.50 & 18.99 $\pm$ 0.03 & 8.20 $\pm$ 0.09  \\
    Online EWC \citep{schwarz2018progress} & 19.96 $\pm$ 0.07 & 33.88 $\pm$ 0.49 & 19.07 $\pm$ 0.13 & 8.38 $\pm$ 0.15 \\
    SI \citep{zenke2017continual} & 19.99 $\pm$ 0.06 & 29.31 $\pm$ 0.62 & 19.14 $\pm$ 0.12 & 9.24 $\pm$ 0.22  \\
    LwF \citep{li2017learning} & 23.85 $\pm$ 0.44 & 22.64 $\pm$ 0.23 & 19.20 $\pm$ 0.30 & 10.71 $\pm$ 0.11  \\
    MAS \citep{aljundi2019task} &  19.50 $\pm$ 0.30 & - & 20.25 $\pm$ 1.54 & 8.44 $\pm$ 0.27  \\
    BGD \citep{zeno2018task} &  19.64 $\pm$ 0.03 & 84.78 $\pm$ 1.30 & - & -  \\
    \bottomrule
    \\
    \\
    \toprule
    \bf With Replay & \bf split MNIST & \bf permuted MNIST & \bf CIFAR-10 & \bf CIFAR-100 \\
    \midrule
    SBC+ER & 90.65 $\pm$ 0.45 & 93.70 $\pm$ 0.09 & 42.07 $\pm$ 0.64 & 28.57 $\pm$ 0.35 \\
    \bf EBM+ER & 91.13 $\pm$ 0.35 & 94.59 $\pm$ 0.09 & \bf 44.76 $\pm$ 0.73 & \bf 34.07 $\pm$ 0.55 \\
    \midrule
    iCaRL \citep{rebuffi2017icarl} & \bf 94.57 $\pm$ 0.11 & \bf  94.85 $\pm$ 0.03 & - & - \\
    DGR \citep{shin2017continual} & 91.30 $\pm$ 0.60 &  92.19 $\pm$ 0.09 &  17.21 $\pm$ 1.88 & 9.22 $\pm$ 0.24 \\
    BI-R \citep{vandeven2020brain} & 94.41 $\pm$ 0.15 & - &  & 21.51 $\pm$ 0.25 \\
    
    GSS-Greedy \citep{aljundi2019gradient} & 84.80 $\pm$ 1.80 & 77.30 $\pm$ 0.50 & 33.56 $\pm$ 1.70 & - \\
    CTN \citep{pham2020contextual} & - & 79.01 $\pm$ 0.65 & - & - \\
    PGMA \citep{hu2019overcoming} & 81.70 $\pm$ (-) & - &  40.47 $\pm$ (-) & - \\
    \bottomrule
    \end{tabular}
    }
    \end{center}
\vspace{-15pt}
\end{table}

\noindent\textbf{Evaluation protocol.}
All experiments are performed according to the class-incremental learning scenario, which is considered as the most natural and also the hardest setting for continual learning \citep{taotopology,he2018exemplar,tao2020few}. Many CL approaches that perform well for task-incremental learning, fail when asked to do class-incremental learning \citep{van2019three}. More details can be found in Appendix~\ref{apx:class-incremental setting}.



\subsubsection{Comparisons with existing methods}
\label{sec:classification_results}
The first comparison of interest is how the performance of the proposed EBMs compares with the performance of the softmax-based classifier (SBC). But we aim higher, and we additionally compare how the proposed EBMs perform relative to methods specifically designed for continual learning. For this, we compare against the methods EWC \citep{kirkpatrick2017overcoming}, Online EWC \citep{schwarz2018progress}, SI \citep{zenke2017continual}, LwF \citep{li2017learning}, MAS \citep{aljundi2019task}, BGD \citep{zeno2018task}, BI-R \citep{vandeven2020brain}, DGR \citep{shin2017continual}, iCaRL \citep{rebuffi2017icarl}, GSS-Greedy \citep{aljundi2019gradient}, 
CTN \citep{pham2020contextual}, and PGMA \citep{hu2019overcoming}.

\textbf{Comparison with methods without replay.}
We first compare EBMs with CL methods without using replay.
In our experiments, we control the baselines and EBMs to have similar model architectures with similar number of model parameters.
For SBC, EWC, Online EWC, Online EWC, LwF on splitMNIST, permuted MNIST, and CIFAR-100, we use the results reported in \citep{van2019three,vandeven2020brain}. For BGD, the results are from \citep{zeno2018task}. For MAS, we use the result from \citep{prabhu12356gdumb,zhang2020self}. We performed the other experiments ourselves using the public available code. The detailed model architectures are listed in Appendix \ref{appendix:model_architectures}.

The \emph{Class-IL} results on four datasets are shown in \tbl{tab:comp_baselines_boundary_aware}.
Experiments ran by ourselves were performed 10 times with different random seeds, with results reported as the mean $\pm$ SEM. Similar training regimes were used for the EBMs and baselines. On split MNIST, permuted MNIST, and CIFAR-10, we trained for 2000 iterations per task. On CIFAR-100, we trained for 5000 iterations per task. 
All experiments used the Adam optimizer with learning rate  $1e^{-4}$.

On these \emph{Class-IL} benchmarks, we observe that EBMs have a significant improvement over SBC, as well as over several softmax-based CL methods that do not rely on an external quota of memory or use generative replay.

Several recent methods~\citep{wortsman2020supermasks,henning2021posterior,yan2021dynamically,douillard2022dytox} also achieve good performance on \emph{Class-IL} without using replay.
However, our main goal is to use EBM, similar to the softmax-based classifier, as a building block for continual learning.
We believe that subsequently optimizing the model architecture, using a larger model, adding additional techniques, or saving previous data, can further improve the performance, but this is not the focus of the paper.



\textbf{Comparison with methods with replay.}
We notice that many of the successful existing methods for \textit{Class-IL} rely on an external quota of memory \citep{rebuffi2017icarl,hayes2020remind} or use generative models \citep{shin2017continual,van2021class,vandeven2020brain,liu2020generative,cong2020gan}. A disadvantage of these methods is that they are relatively expensive in terms of memory and/or computation.
Thus it is not fair to compare the basic version of EBMs to methods using replay.

Importantly, however, EBMs are flexible enough to be combined with existing continual learning approaches to further improve the performance.
In \tbl{tab:comp_baselines_boundary_aware}, we show the result of combining EBMs with exact replay, \ie \textbf{EBM+ER}. 
As EBM+ER uses previous data during training, for fair comparisons, we design another baseline that combines SBC with exact replay, \ie SBC+ER.
See Appendix \ref{apx:replay_details} for the implementation details of EBM+ER and SBC+ER.
Combined with replay, both SBC and EBM have improvements. EBMs still outperform SBC, especially on more challenging datasets, \eg CIFAR-10 and CIFAR-100. 
Interestingly, on CIFAR-100, we find that EBMs without using replay (EBM: 30.28\%) perform better than SBC with using replay (SBC+ER: 28.57\%).
Moreover, the performance of EBM+ER is comparable to the performance of established replay-based methods such as iCaRL, DGR, and BI-R.



Our results indicate that the proposed EBM formulation provides a promising building block for tackling CL problems.
Other approaches, such as replay-based approaches, can build on top of EBMs.


\begin{table}[t]
  \small
  \caption{\small{Class-IL results of EBM and baselines using our contrastive divergence training objective and their original one.}}
  \vspace{-5pt}
  \label{tab:new_train_strategy}
  \centering
    \setlength{\tabcolsep}{4pt}
    \resizebox{0.8\textwidth}{!}{
    \begin{tabular}{llccccc} 
        \toprule
        \bf Dataset & \bf Method & \bf CLS & \bf EWC & \bf Online EWC & \bf SI & \bf EBM  \\
        \midrule
        \multirow{2}*{\bf split MNIST} 
        & Original Objective & 19.90 $\pm$ 0.02  & 20.01 $\pm$ 0.06 & 19.96 $\pm$ 0.07 & 19.99 $\pm$ 0.06 & -  \\
        & CD Objective & 44.98 $\pm$ 0.05 & 50.68 $\pm$ 0.04 & 50.99 $\pm$ 0.03 & 49.44 $\pm$ 0.03 & \bf 53.12 $\pm$ 0.04 \\
        \midrule
        \multirow{2}*{\bf CIFAR-10} 
        & Original Objective & 19.06 $\pm$ 0.01 & 18.99 $\pm$ 0.01 & 19.07 $\pm$ 0.01 & 19.14 $\pm$ 0.02 & - \\
        & CD Objective & 19.22 $\pm$ 0.02 & 36.51 $\pm$ 0.03 & 36.16 $\pm$ 0.02 & 35.12 $\pm$ 0.02 & \bf 38.84 $\pm$ 0.01 \\
        \bottomrule
    \end{tabular}
    }
\end{table}


\subsubsection{Effect of Contrastive divergence training objective and label conditioning}

\noindent\textbf{Contrastive divergence training objective on existing approaches.}
The proposed contrastive divergence training objective is simple and can also be directly applied to existing CL approaches.
We test this by modifying the training objective of baseline models to that of our proposed contrastive divergence objective, which computes the softmax normalization only over the ground truth class $y$ and a negative class $i$ sampled from the current training batch: $\mathcal{L}_{\text{SBC\_CD}}(\rvx,y) = -[f_{\boldsymbol{\theta}}(\rvx)]_{y} + [f_{\boldsymbol{\theta}}(\rvx)]_{i}$. 
We only modify their training objective without changing their model architectures. 
In \tbl{tab:new_train_strategy}, we find that our proposed training objective (CD Objective) significantly improves the performance of different CL methods. 
This is because the contrastive divergence objective does not suppress the probability of old classes when improving the probability of new classes.
Our training objective is simple to implement on existing CL approaches.




\begin{figure*}[t]
\centering
\hfill
\begin{minipage}{0.29\textwidth}
    \centering
    \small
    \captionof{table}{\small{Performance of EBM on CIFAR-100 with different strategies for selecting the negative samples.}}
      \label{tab:neg_sam}
      \centering
      \scalebox{1}{
      \begin{tabular}{lccc}
        \toprule
        \bf Dataset & \bf CIFAR-100 \\
        \midrule
        All Neg Seen & 8.07 $\pm$ 0.10 \\
        All Neg Batch & 29.03 $\pm$ 0.53 \\
        \bf 1 Neg Batch & \bf 30.28 $\pm$ \bf 0.28 \\
        \bottomrule
      \end{tabular}
      }
    \end{minipage}%
\hfill
\begin{minipage}{0.65\textwidth}
    \centering
    \small
    \includegraphics[width=1\linewidth]{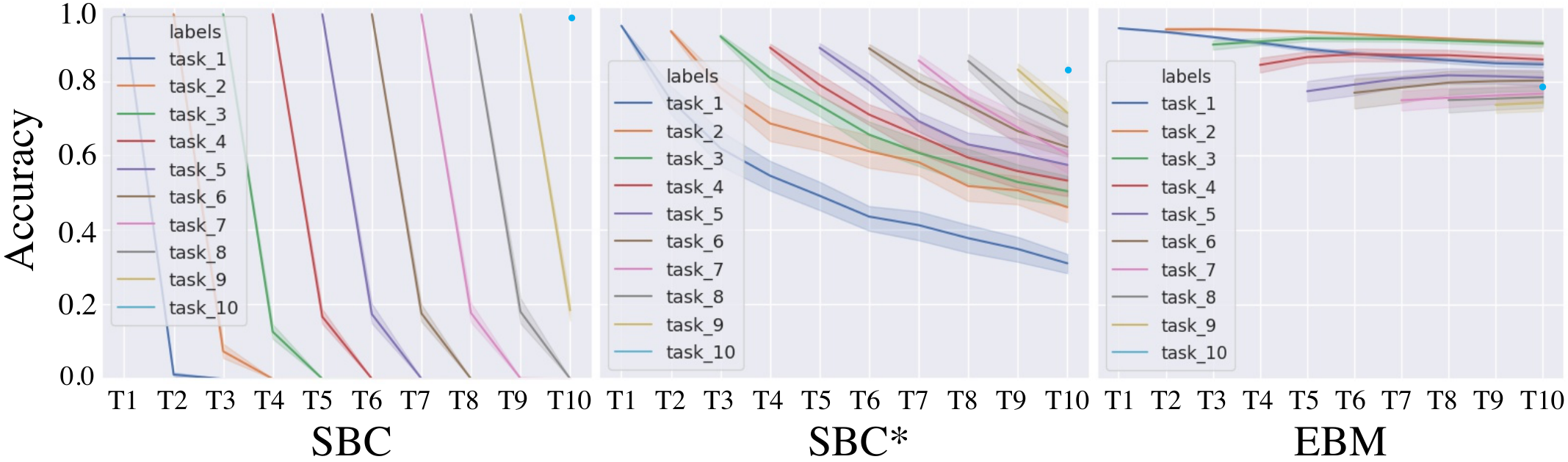}
    \vspace{-20pt}
    \caption{\small{\emph{Class-IL} testing accuracy of SBC, SBC using our training objective (SBC*), and EBMs on each task on the permuted MNIST dataset.}}
    \label{fig:training_curve}
\end{minipage}
\vspace{-10pt}
\end{figure*}

\noindent\textbf{Effect of training objectives.}
We conduct an experiment on the CIFAR-100 dataset  to investigate how different EBM training objectives influence the CL results.
We compare three different strategies for selecting the negative samples as described in \sect{sec:ebm_training_objective}.
%
The first strategy uses all seen classes so far as negative labels (\textbf{All Neg Seen}), which is most similar to the way that the traditional classifier is optimized. The second one takes all the classes in the current batch as negative labels (\textbf{All Neg Batch}). The last one --- the contrastive divergence objective we proposed in \eqn{eqn:cd_loss} --- randomly selects one class from the current batch as the negative (\textbf{1 Neg Batch}). 
In \tbl{tab:neg_sam}, we find that using only one negative sample generates the best result, and using negatives sampled from classes in the current batch is better than from all seen classes.
Since our EBM training objective aims at improving the energy of negative samples while decreasing the energy of positive ones, sampling negatives from the current batch has less interference with previous classes than sampling from all seen classes.
The proposed contrastive divergence training objective uses a single negative and causes the minimum suppression on negative samples.



\noindent\textbf{Effect of label conditioning.}
Next, we test whether the label conditioning in our EBMs is important for their performance. 
As mentioned in \tbl{tab:new_train_strategy}, we modify baselines using our training objective.
The only difference is that EBMs have the label conditioning architecture while baselines do not.
EBMs outperform the modified baselines, \eg Online EWC --- the best baselines with the contrastive divergence training objective --- is $50.99\%$ while EBM is $53.12\%$ on split MNIST. 
This implies that the label conditioning architecture mitigates catastrophic forgetting in EBMs.
%

We further show the testing accuracy of each task as the training progresses in \fig{fig:training_curve}.
We compare the standard classifier (SBC), classifier using the contrastive divergence training objective (SBC*), and our EBMs on the permuted MNIST dataset. 
We find that the accuracy of old tasks in SBC drop sharply when learning new tasks, while the contrastive divergence training objective used in SBC* is better. The curve on EBMs drops even slower than SBC*. As both SBC* and EBMs use the contrastive divergence training objective, the good performance of EBMs implies the effectiveness of the label conditioning architecture in EBMs.


To summarize, we show that the strong performance of our EBMs is due to both the contrastive divergence training objective and the label conditioning architecture. Moreover, these results indicate that surprisingly, and counterintuitively, directly optimizing the softmax-based architecture with the cross-entropy loss used by existing approaches may not be the best way to approach continual learning.

\vspace{-5pt}
\subsubsection{Comparisons of different model architectures}
\label{sec:model_architectures}
EBMs allow flexibility in integrating data information and label information in the energy function.
To investigate where and how to combine the information from data $\rvx$ and label $y$, we conduct a series of experiments on CIFAR-10.
\tbl{tab:diff_net} shows four model architectures (V1-V4) that combine $\rvx$ and $y$ in the early, middle, and late stages, respectively (see Appendix \ref{appendix:model_architectures} for more details). We find that combining $\rvx$ and $y$ in the late stage (V4) performs the best.
We note that instead of learning a feature embedding of label $y$, we can use a fixed projection matrix which is randomly sampled from the uniform distribution $\mathcal{U}(0, 1)$. Even using this fixed random projection can already generates better results than most baselines without replay in \tbl{tab:comp_baselines_boundary_aware}. Note further that the number of trainable parameters in the ``Fix" setting is much lower than that of the baselines. 
We notice that the standard classifiers have to change their model
architecture by adding new class heads to the softmax output
layer when dealing with new classes. In contrast, the EBMs ``Fix" setting do not need to modify the model architecture or resize the network when adding new classes.
Using a learned feature embedding of $y$ can further improve the result. 
We apply different normalization methods over the feature channel of $y$. 
We find that Softmax ({End Softmax (V4)}) is better than the L2 normalization ({End Norm2 (V4)}) and no normalization ({End (V4)}). 
\begin{table}[t]
\centering
\hfill
    \caption{\small{Performance of EBM with different label conditioning architectures on the CIFAR-10 dataset.}}
    \label{tab:diff_net}
    \begin{center}
    \begin{small}
    \scalebox{0.99}{
    \begin{tabular}{lc|lc}
    \toprule
    \multicolumn{2}{c}{\bf Model architectures} & \multicolumn{2}{c}{\bf Normalization types} \\
    \midrule
    Beginning (V1) & 13.69 $\pm$ 1.12 & End Fix (V4) & 34.30 $\pm$ 1.03 \\ 
    Middle (V2) & 20.16 $\pm$ 1.05 & End Fix Norm2 (V4) & 33.91 $\pm$ 1.13 \\ 
    Middle (V3) & 18.36 $\pm$ 0.97 & End Fix Softmax (V4) & 35.97 $\pm$ 1.09 \\
    \bf End (V4) & 38.13 $\pm$ 0.59 & End Norm2 (V4) & 37.23 $\pm$ 1.20 \\
    && \bf End Softmax (V4) & \bf 38.84 $\pm$ 1.08 \\
    \bottomrule
    \end{tabular}
    }
    \end{small}
    \end{center}
\vspace{-15pt}
\end{table}



\subsubsection{Qualitative analysis}
\label{sec:qualitative}
To better understand why EBMs suffer less from catastrophic forgetting, we qualitatively compare our EBMs and SBC in this part.
We provide additional analyses of CL performance of SBC and EBMs in Appendix~\ref{sec:additional_results} and Appendix~\ref{sec:additional_analyses}.

\begin{figure*}[t]
\centering
\begin{minipage}{0.55\textwidth}
    \centering
    \includegraphics[width=1\textwidth]{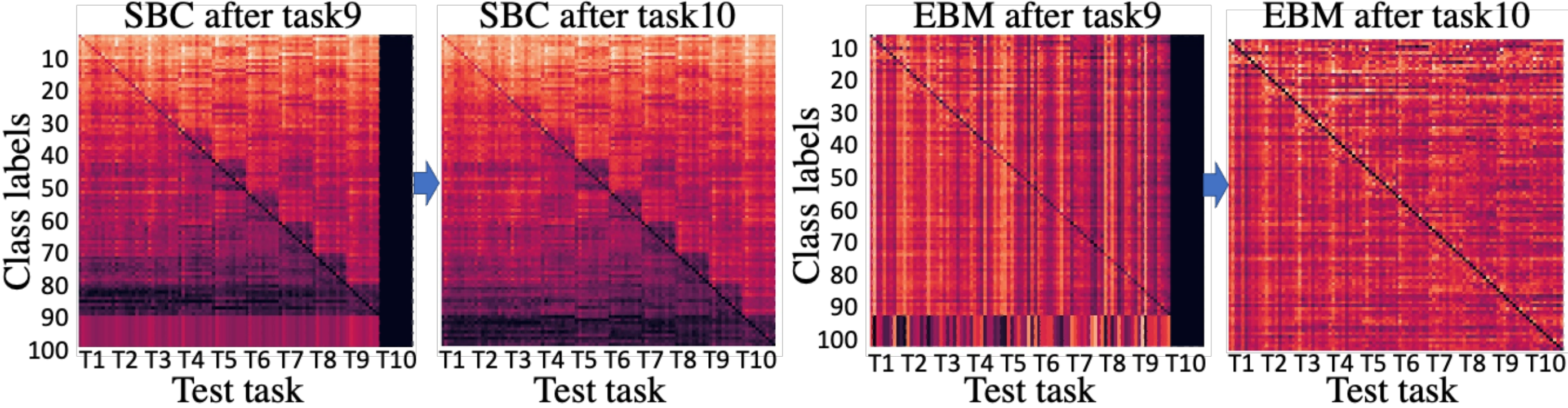}
    \vspace{-10pt}
    \caption{\small{Energy landmaps of SBC and EBMs after training on task $T_9$ and $T_{10}$ on permuted MNIST. The darker the diagonal is, the better the model is in preventing forgetting previous tasks.}}
    \label{fig:energy_landmap}
    \end{minipage}%
    \hfill
    \begin{minipage}{0.43\textwidth}
    \centering
    \includegraphics[width=1\textwidth]{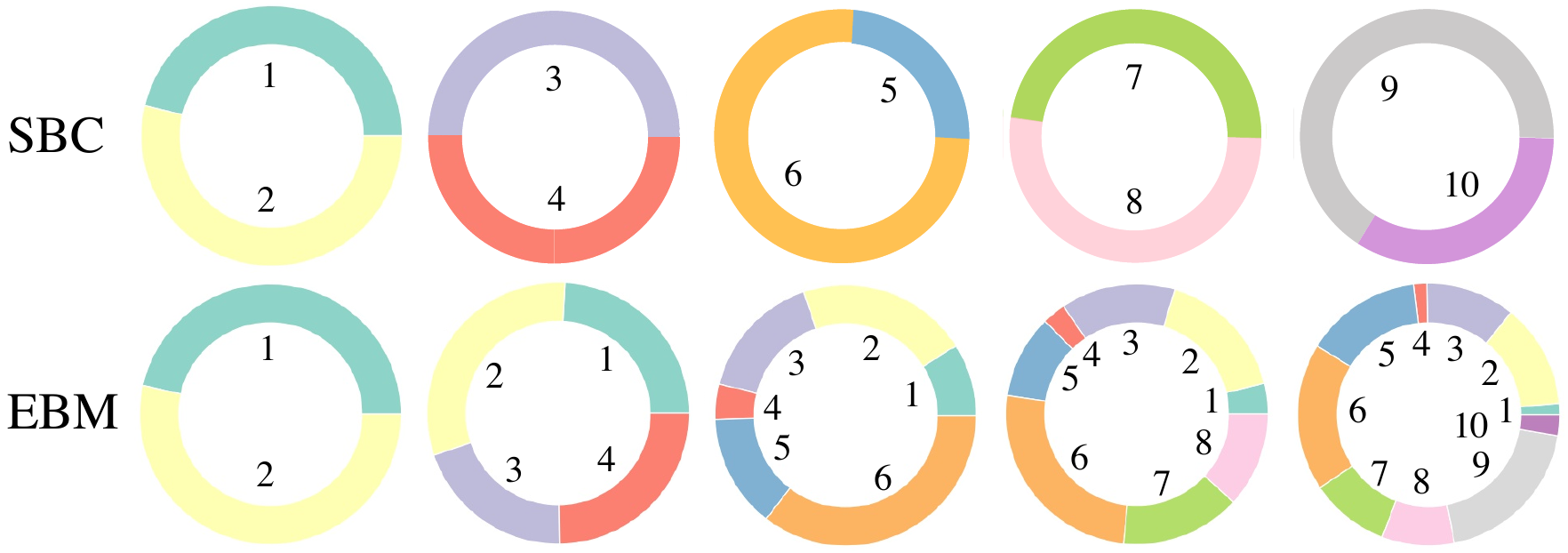}
    \vspace{-15pt}
    \caption{\small{Predicted label distribution after learning each task on split MNIST. SBC only predicts classes from the current task, while EBM predicts classes from all seen classes.}
    }
    \label{fig:new_label_distributation}
\end{minipage}
\end{figure*}

\noindent\textbf{Energy landscape.}
In \fig{fig:energy_landmap}, we show the energy landscapes after training on task 9 and task 10 of the permuted MNIST dataset. For SBC, the energy is given by the negative of the predicted probability. Each datapoint has 100 energy values (EBM) or probabilities (SBC) corresponding to the 100 labels in the dataset. For each datapoint, these values are normalized over all 100 classes. 
Dark elements on the diagonal indicate correct predictions.
After training on task $T_9$, SBC assigns high probabilities to classes from $T_9$ (80-90) for almost all the data from $T_1$ to $T_9$. After learning task $T_{10}$, the highest probabilities shift to classes from task $T_{10}$ (90-100). 
SBC tends to assign high probabilities to new classes for both old and new data, indicating forgetting.
In contrast, EBM has low energies across the diagonal, which means that after training on new tasks, EBM still assigns low energies to the true labels of data from previous tasks.
This shows that EBM is better than SBC at learning new tasks without catastrophically forgetting the old tasks.

\noindent\textbf{Predicted class distribution.}
\label{sec:label_distributation}
In \fig{fig:new_label_distributation}, for the split MNIST dataset, we plot the proportional distribution of predicted classes. Only data from the tasks seen so far was used for this figure. Taking the second panel in the first row as an example, it shows the distribution of predicted labels on test data from the first two tasks after finishing training on the second task. This means that so far the model has seen four classes: \{1,2,3,4\}. Since the number of test images from each class are similar, the ground truth proportional distribution should be uniform over those four classes. After training on the first task, the predictions of SBC are indeed roughly uniformly distributed over the first two classes (first panel). However, after learning new tasks, SBC only predicts classes from the most recent task, indicating that SBC fails to correctly memorize classes from previous tasks. In contrast, the predictions of EBM are substantially more uniformly distributed over all classes seen so far.

\subsection{Experiments on Boundary-Free Setting}
\label{exp:boundary-free}
When applying continual learning in real life, boundaries are not usually well defined between different tasks.
However, most existing CL methods rely on the presence of sharp, known boundaries between tasks to determine when to consolidate the knowledge. 
We show that EBMs are able to flexibly perform CL across different setups, and perform well on the \emph{boundary-free} setting as well.

\subsubsection{Datasets and evalution protocols}
For the \emph{boundary-free} setting, we use the same datasets as the \emph{boundary-aware} setting in \sect{exp:boundary-aware}. 
We use the code of ``continuous
task-agnostic learning" proposed by \citep{zeno2018task} to generate a continually changing data stream during training. In this setting, the frequency of each subsequent class increases and decreases linearly. 
See Appendix \ref{apx:boundary-free setting} for more details of the \emph{boundary-free} setting. All experiments are again performed in a class-incremental setup.

\begin{table}
\small
\caption{\small Evaluation of class-incremental learning performance on the \emph{boundary-free} setting. 
Each experiment is performed 5 times with different random seeds, average test accuracy is reported as the mean $\pm$ SEM over these runs. 
}
\vspace{-10pt}
\label{tab:comp_baselines_boundary_free}
\begin{center}
\scalebox{0.87}{
\begin{tabular}{lcccc}
\toprule
\bf Method & \bf split MNIST & \bf permuted MNIST & \bf CIFAR-10 & \bf CIFAR-100 \\
\midrule
SBC & 24.03 $\pm$ 0.59 & 21.42 $\pm$ 0.11 & 23.30 $\pm$ 0.81 & 9.85 $\pm$ 0.02 \\
\bf EBM & \bf 81.78 $\pm$ 1.22 & \bf 92.35 $\pm$ 0.11 & \bf 49.47 $\pm$ 1.25 & \bf 34.39 $\pm$ 0.24 \\
\midrule
Online EWC \citep{schwarz2018progress} & 39.62 $\pm$ 0.14 & 41.37 $\pm$ 0.04 & 22.53 $\pm$ 0.41 & 9.57 $\pm$ 0.02 \\
SI \citep{zenke2017continual} & 28.79 $\pm$ 0.24 & 35.71 $\pm$ 0.11 & 26.26 $\pm$ 0.72 & 10.42 $\pm$ 0.01 \\
BGD \citep{zeno2018task} & 21.65 $\pm$ 1.15 & 26.15 $\pm$ 0.22 & 17.03 $\pm$ 0.82 & 8.50 $\pm$ 0.02 \\
\bottomrule
\end{tabular}
}
\end{center}
\vspace{-15pt}
\end{table}


\subsubsection{Comparison with existing methods}
\label{sec:classification_results_boundary_free}
In this setting, we restrict our comparison to methods that do not use replay. Most of the non-replay methods that were compared in Table~\ref{tab:comp_baselines_boundary_aware} cannot be applied to the \emph{boundary-free} setting without modification, because these methods rely on known task boundaries to perform certain consolidation operations (e.g., to update the parameter regularization term). A relatively straight-forward adaptation of these methods is to perform their consolidation operation after every mini-batch step instead of after each task (see also \citep{zeno2018task}), although this adaptation is impractical for some algorithms (e.g., EWC) because of large computational complexity. We managed to run the Online EWC and SI baselines in this way, while the BGD baseline is designed to be suitable for the \emph{boundary-free} setting.

We further note that there are continual learning methods that might be applicable to settings in which task boundaries are unknown (i.e., a \emph{boundary-agnostic} setting), but these methods are not necessarily also applicable to the \emph{boundary-free} setting, because often these methods still require that there are underlying task boundaries. For example, the method proposed by \citep{wortsman2020supermasks} is able to infer task boundaries if they are not given (i.e., \emph{boundary-agnostic}), but this method cannot deal with the setting in which there are no underlying boundaries at all (i.e., \emph{boundary-free}).

The results on the \emph{boundary-free} setting are shown in \tbl{tab:comp_baselines_boundary_free}. All compared methods used similar model architectures as in the \emph{boundary-aware} setting. Each  experiment  was  performed $5$ times  with different random seeds, the results are reported as the mean $\pm$ SEM over these runs.
We observe that EBMs have a significant improvement on all the datasets.
The experiments show that EBMs have good generalization ability for different continual learning problems as EBMs can naturally handle data streams with and without task boundaries.

\section{Discussion}
\label{sect:conclusion}
In this paper, we show that energy-based models are a promising class of models in a variety of different continual learning settings. 
We demonstrate that EBMs exhibit many desirable characteristics to prevent catastrophic forgetting in continual learning, and we
experimentally show that EBMs obtain strong performance on the challenging class-incremental learning scenario on multiple benchmarks, both on the boundary-aware and boundary-free settings.
One drawback of the current EBM formulation is that we need to compute the energy between a data point and each class during inference.
However, in the actual implementation, we found that the difference between the softmax-based classifier (SBC) and our EBM is minor. 
Another potential limitation of this work is that we only focus on evaluating the average accuracy, and mostly only after training has finished. It might be interesting to also focus on other evaluation metrics, such as backward and forward transfer that evaluates models’ ability to transfer knowledge across tasks \citep{lopez2017gradient,vogelstein2020omnidirectional}, or to perform continual evaluation \citep{delange2022cleval}.
We do provide several additional results and analyses in the appendix.

\section*{Acknowledgements}
This work has been partly supported by the Lifelong Learning Machines (L2M) program of the
Defence Advanced Research Projects Agency (DARPA) via contract number HR0011-18-2-0025.

\bibliography{collas2022_conference}
\bibliographystyle{collas2022_conference}

\clearpage

\appendix

\noindent\textbf{\Large{Appendices}}
\vspace{5pt}


In this appendix, we provide more details of the experiment setup in \sect{sec:details}. We show more experimental results of our EBMs and baselines in \sect{sec:additional_results}. 
\sect{appendix:model_architectures} lists the model architecture details of the proposed EBMs and baselines on difference datasets. \sect{apx:replay_details} has more experimental details of EBMs using replay. We provide additional analysis of why EBM can mitigate the catastrophic forgetting problem in continual learning in \sect{sec:additional_analyses}.
\sect{apx:alternative_training_objective_appendix} describes the alternative EBM training objective mentioned in the main paper \sect{sec:method_ebm_classification_cl}.

\section{More Experiment Setup Details}
\label{sec:details}

\subsection{More details about the class-incremental setting}
\label{apx:class-incremental setting}
In the main paper \sect{sec:continual_settings}, we have talked about the task-incremental and class-incremental settings. In this part, we provide more details of the class-incremental setting.

Class-incremental learning is a standard continual learning setting and has been used in many existing continual learning papers. 
In the class-incremental learning, the model is sequentially trained on each task. During testing, the model needs to predict the correct class from all classes.

\textbf{Training:} Taking the split MNIST dataset under the boundary-aware setting as an example, task1 has images from two classes [0, 1]; task2 has images from two classes [2, 3]; ...; and task5 has images from classes [8, 9]. The model is first trained on task1, and then task2, until the final task. Note that the model is trained sequentially, so when training the final task, the model only sees images and classes in the final task, and thus it tends to forget previous tasks which is called catastrophic forgetting.

\textbf{Testing:} In the class-incremental learning, given an image from a task, the model needs to predict its class label from all seen classes (e.g. 10 classes in split MNIST). Note that this is challenging as 1) during training, the model only needs to predict the class label of an image from classes in the current task (e.g. 2 classes in split MNIST); 2) after the model is trained on the final task, it is more likely to predict classes from the final task (e.g. classes [8, 9] in split MNIST), even the input image is from previous tasks. Please see \fig{fig:energy_landmap} and \fig{fig:new_label_distributation} in the main paper for a better understanding.

\subsection{More details about the boundary-free and boundary-aware settings}
\label{apx:boundary-free setting}
In the main paper \sect{sec:continual_settings}, we have talked about the boundary-free and boundary-aware settings. In this part, we provide more details of the less frequently used boundary-free setting.

Boundary-free and boundary-aware are two different settings of how the data is provided during training. We follow the general implementation of \citep{zeno2018task} for the boundary-free setting. In this setting, models learn in a streaming fashion and the data distributions gradually change over time. Taking the split MNIST dataset as an example, it has 10 classes. During training, the percentage of `1s' in each training batch gradually decreases while the percentage of `2s' increases, and then the percentage of `2s' in each training batch gradually decreases while the percentage of `3s' increases, and so on. The model can observe data from different classes during training and there is no a clear task boundary. It might also be useful to have a look at Figure 2 of the paper \citep{zeno2018task} for a better intuitive understanding of the boundary-free setting.

In our contrastive divergence training objective, we randomly select 1 negative sample from the current batch. If there are multiple negative samples, we just randomly select 1 from them.
In \tbl{tab:neg_sam} of the main paper, our contrastive divergence training objective (1 Neg Batch=30.28) is better than using all the negative classes in the current batch (All Neg Batch=29.03) and is much better than using all the negative classes seen so far (All Neg Seen=8.07).

In the contrastive divergence training objective, the positive and negative classes are sampled from the current batch regardless of how the batch data is provided. The batch data can be sampled based on either the boundary-free setting or the boundary-aware setting. For example, In the boundary-aware setting, the first batch may only contain classes (1,2) from the first task, the second batch may only contain classes (3,4) from the second task, the third batch may only have classes (5,6) from the third task, and so on. There are clear task boundaries. In the boundary-free setting, the first batch may contain classes (1), later batches may contain classes (1,2), and then classes (2,3), and so on. There are no task boundaries.

The contrastive divergence loss allows our method to be trained without requiring knowledge of the task boundaries, and thus it can naturally handle both the boundary-aware and boundary-free settings.
In contrast, many existing CL approaches, such as EWC, LwF, are not directly applicable to the boundary-free setting as they require the task boundaries to decide when to perform certain consolidation steps.


\begin{table}[t]
\small
    \caption{\small Comparison of our EBM with baselines on different variants of the split CIFAR-100 protocol. 
    Results of the \emph{Class-IL} performance on the \emph{boundary-aware} setting are reported.
    }
    \vspace{-10pt}
    \setlength{\tabcolsep}{4pt}
    \label{tab:number_of_tasks}
    \begin{center}
    \begin{small}
    \scalebox{0.85}{
    \begin{tabular}{lcccc}
    \toprule
    \multirow{2}{*}{\bf Method} & \multicolumn{4}{c}{\bf CIFAR-100 split up into:} \\
    & \bf 5 tasks & \bf 10 tasks & \bf 20 tasks & \bf 50 tasks \\
    \midrule
    SBC & 14.74 $\pm$ 0.20 & 8.18 $\pm$ 0.10 & 4.46 $\pm$ 0.03 & 1.91 $\pm$ 0.02 \\
    \bf EBM & \bf 34.88 $\pm$ 0.14 & \bf 30.28 $\pm$ 0.28 & \bf 25.04 $\pm$ 0.33 & \bf 13.60 $\pm$ 0.50  \\
    \midrule
    EWC \citep{kirkpatrick2017overcoming} & 14.78 $\pm$ 0.21 & 8.20 $\pm$ 0.09 & 4.46 $\pm$ 0.03 & 1.91 $\pm$ 0.02  \\
    SI \citep{zenke2017continual} & 14.07 $\pm$ 0.24 & 9.24 $\pm$ 0.22 & 4.37 $\pm$ 0.04 & 1.88 $\pm$ 0.03  \\
    LwF \citep{li2017learning} & 25.75 $\pm$ 0.14 & 10.71 $\pm$ 0.11 & 12.18 $\pm$ 0.16 & 7.68 $\pm$ 0.16  \\
    \bottomrule
    \end{tabular}
    }
    \end{small}
    \end{center}
\vspace{-10pt}
\end{table}

\section{Additional Results}
\label{sec:additional_results}

Extending the results presented in the main paper, here we further compare EBMs with the baseline models by providing additional experiments of their performance. 
We first investigate the effect of different numbers of tasks in \sect{sec:number_of_tasks}.
We then evaluate the computational complexity for inference in \sect{app_sec:inference} and show the confusion matrices between the ground truth labels and model predictions in \sect{sec:confusion_matrix}.

\begin{figure*}[t]
\begin{center}
\includegraphics[width=0.8\textwidth]{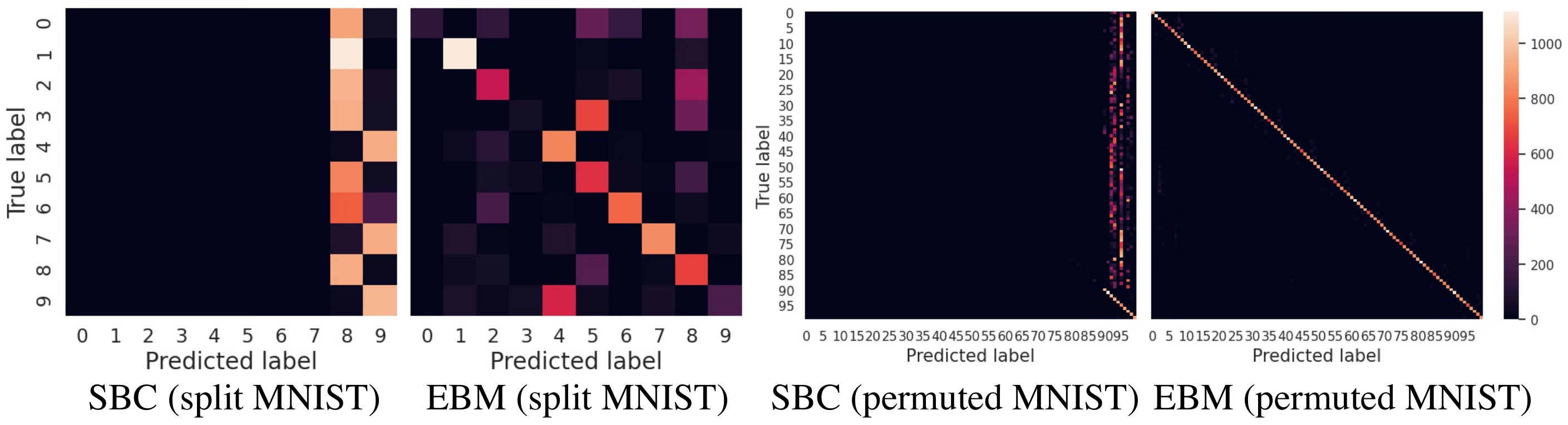}
\end{center}
\vspace{-10pt}
\caption{\small{Confusion matrices between ground truth labels and predicted labels at the end of learning on split MNIST (left) and permuted MNIST (right). The lighter the diagonal is, the more accurate the predictions are.}}
\label{fig:new_confusion_matrix}
\vspace{-5pt}
\end{figure*}

\subsection{Effect of different numbers of tasks}
\label{sec:number_of_tasks}
To test the generality of our proposed EBMs, we repeat the boundary-aware experiments on CIFAR-100 for different number of classes per task in Table~\ref{tab:number_of_tasks}. In the main paper Table~\ref{tab:comp_baselines_boundary_aware}, the CIFAR-100 dataset was split into 10 tasks, resulting in 10 classes per task. Here we additionally split CIFAR-100 into 5 tasks (i.e., 20 classes per task), 20 tasks (i.e., 5 classes per task), and 50 tasks (i.e., 2 classes per task). Our EBM substantially outperforms the baselines on all settings.

\subsection{Inference computational complexity}
\label{app_sec:inference}
One drawback of the current EBM formulation is that we need to compute the energy between a data point and each class during inference.
However, in the actual implementation, we found that the difference between the softmax-based classifier (SBC) and our EBM is minor. 

We test the inference time needed for a single forward sweep of the softmax-based classifier (SBC) and our EBM on the split MNIST and CIFAR-10 datasets. We use TITAN Xp 12G GPUs for testing. We evaluate every single model on 1 single GPU. Each experiment is run 5 times with different random seeds.

On the split MNIST dataset, the average inference time of SBC is 0.00022s while EBM is 0.00027s. On CIFAR-10, the average inference time of SBC is 0.00059s while EBM is 0.00063s. The difference between SBC and EBM is less than 0.0001s for a single forward pass.

\subsection{Class confusion matrix at the end of learning}
\label{sec:confusion_matrix}
We show confusion matrices for EBMs and SBC. A confusion matrix illustrates the relationship between the ground truth labels and the predicted labels. \fig{fig:new_confusion_matrix} shows the confusion matrices after training on all the tasks on the split MNIST dataset and permuted MNIST dataset. The standard classifier tends to only predict the classes from the last task (class 8, 9 on the split MNIST dataset and classes 90-100 on the permuted MNIST dataset). The EBMs on the other hand have high values along the diagonal, which indicates that the predicted results match the ground truth labels for all the sequentially learned tasks.

The poor performance of SBC might be caused by an imbalance between bias and norms of vectors biased toward the last classes in a fully-connected layer.
\citep{wu2019large} and \citep{hou2019learning} correct the classification result by reserving the samples for old classes and/or training extra correction layers. \citep{wu2019large} train a bias correction layer using the data from both the old classes and the new classes. \citep{hou2019learning} also uses old data to rebalance the training process. In contrast, EBM can predict the classes correctly without storing previous data and without training extra layers to correct the bias.
\citep{lesort2021continual} investigates different sources of performance decrease for the output layer and proposes three different methods to address the catastrophic forgetting in the output layer: a simplified weight normalization layer, two masking strategies, and an alternative to Nearest Mean Classifier using median vectors. Some techniques may also be applied to EBM, such as weight normalization.

\section{Model architectures}
\label{appendix:model_architectures}
In this section, we provide details of the model architectures used on different datasets. 

Images from the split MNIST and permuted MNIST datasets are grey-scale images. The baseline models for these datasets, similar as in
\citep{van2019three}, consist of several fully-connected layers.
We use model architectures with similar number of parameters for EBMs. 
The model architectures of EBMs on the split MNIST dataset and permuted MNIST dataset are the same, but have different input and output dimensions and hidden sizes.
The model architectures of EBMs and baseline models on the split MNIST dataset are shown in \tbl{tab:mnist}.
The model architectures of EBMs and baseline models on the permuted MNIST dataset are shown in \tbl{tab:perm}.

Images from the CIFAR-10 and CIFAR-100 datasets are RGB images. For CIFAR-10, we use a small convolutional network for both the baseline models and EBMs.
We investigate different architectures to search for the effective label conditioning on EBMs training as described in the main paper \sect{sec:model_architectures}.
The model architectures of EBMs and baseline models on the CIFAR-10 dataset are shown in \tbl{tab:ebm_cifar10}.
The model architectures used on the CIFAR-100 dataset are shown in \tbl{tab:cifar100}.


\begin{table*}[t]
  \caption{\small{The model architectures used on split MNIST. $h=400$.}}
  \vspace{-5pt}
  \small
  \label{tab:mnist}
  \begin{minipage}[b]{0.4\textwidth}
  \subcaption{\small The architecture of EBMs.}
  \vspace{-5pt}
  \label{tab:ebm_mnist}
  \centering
  \begin{tabular}{l}
    \toprule
    x = FC(784, h) (x) \\
    \midrule 
    y = Embedding (y) \\
    \midrule
    x = x * Norm2 (y) + x \\
    \midrule 
    x = ReLU (x) \\
    \midrule 
    out = FC(h, 1) (x) \\
    \bottomrule
  \end{tabular}
  \end{minipage}
  \hfill
  \begin{minipage}[b]{0.4\textwidth}
  \subcaption{\small The architecture of baseline models.}
  \vspace{-5pt}
  \label{tab:cls_mnist}
  \centering
  \begin{tabular}{l}
    \toprule
    x = FC(784, h) (x) \\
    \midrule 
    x = ReLU (x) \\
    \midrule 
    x = FC(h, h) (x) \\
    \midrule 
    x = ReLU (x) \\
    \midrule 
    out = FC(h, 10) (x) \\
    \bottomrule
  \end{tabular}
  \end{minipage}
\end{table*}

\begin{table*}[t]
  \caption{\small{The model architectures used on permuted MNIST. $h=1000$.}}
  \vspace{-8pt}
  \small
  \label{tab:perm}
  \begin{minipage}[b]{0.4\textwidth}
  \subcaption{\small The architecture of EBMs.}
  \vspace{-5pt}
  \label{tab:ebm_perm}
  \centering
  \begin{tabular}{l}
    \toprule
    x = FC(1024, h) (x) \\
    \midrule 
    y = Embedding (y) \\
    \midrule
    x = x * Norm2 (y) + x \\
    \midrule 
    x = ReLU (x) \\
    \midrule 
    out = FC(h, 1) (x) \\
    \bottomrule
  \end{tabular}
  \end{minipage}
  \hfill
  \begin{minipage}[b]{0.4\textwidth}
  \vspace{-5pt}
  \subcaption{\small The architecture of baseline models.}
  \vspace{-5pt}
  \label{tab:cls_perm}
  \centering
  \begin{tabular}{l}
    \toprule
    x = FC(1024, h) (x) \\
    \midrule 
    x = ReLU (x) \\
    \midrule 
    x = FC(h, h) (x) \\
    \midrule 
    x = ReLU (x) \\
    \midrule 
    out = FC(h, 100) (x) \\
    \bottomrule
  \end{tabular}
  \end{minipage}
\vspace{-10pt}
\end{table*}

\begin{table*}[t]
\vspace{-8pt}
  \caption{\small{The model architectures used on the CIFAR-10 dataset.}}
  \small
  \label{tab:ebm_cifar10}
  \begin{minipage}[b]{0.3\textwidth}
  \subcaption{\small EBM: \textbf{Beginning (V1)}}
  \vspace{-5pt}
  \label{tab:ebm_cifar10_v1}
  \centering
  \begin{tabular}{l}
    \toprule
    Input: x, y \\
    \midrule 
    y = Embedding(N, 3) (y) \\
    \midrule 
    y = Softmax(dim=-1) (y) \\
    \midrule 
    y = y * y.shape[-1] \\
    \midrule 
    x = x * y \\
    \midrule 
    x = Conv2d(3$\times$3, 3, 32) (x) \\
    \midrule 
    x = ReLU (x) \\
    \midrule
    x = Conv2d(3$\times$3, 32, 32) (x) \\
    \midrule
    x = ReLU (x) \\
    \midrule
    x = Maxpool(2, 2) (x) \\
    \midrule
    x = Conv2d(3$\times$3, 32, 64) (x) \\
    \midrule
    x = ReLU (x) \\
    \midrule
    x = Conv2d(3$\times$3, 64, 64) (x) \\
    \midrule
    x = ReLU (x) \\
    \midrule
    x = Maxpool(2, 2) (x) \\
    \midrule
    x = FC(2304, 1024) (x) \\
    \midrule
    x = ReLU(x) \\
    \midrule
    out = FC(1024, 1) (x) \\
    \bottomrule
  \end{tabular}
  \end{minipage}
  \hfill
  \begin{minipage}[b]{0.3\textwidth}
  \subcaption{\small EBM: \textbf{Middle (V2)}}
  \vspace{-5pt}
  \label{tab:ebm_cifar10_v2}
  \centering
  \begin{tabular}{l}
    \toprule
    Input: x, y \\
    \midrule 
    x = Conv2d(3$\times$3, 3, 32) (x) \\
    \midrule 
    x = ReLU (x) \\
    \midrule
    x = Conv2d(3$\times$3, 32, 32) (x) \\
    \midrule
    x = ReLU (x) \\
    \midrule
    y = Embedding(N, 32) (y) \\
    \midrule 
    y = Softmax(dim=-1) (y) \\
    \midrule 
    y = y * y.shape[-1] \\
    \midrule 
    x = x * y \\
    \midrule 
    x = Maxpool(2, 2) (x) \\
    \midrule
    x = Conv2d(3$\times$3, 32, 64) (x) \\
    \midrule
    x = ReLU (x) \\
    \midrule
    x = Conv2d(3$\times$3, 64, 64) (x) \\
    \midrule
    x = ReLU (x) \\
    \midrule
    x = Maxpool(2, 2) (x) \\
    \midrule
    x = FC(2304, 1024) (x) \\
    \midrule
    x = ReLU(x) \\
    \midrule
    out = FC(1024, 1) (x) \\
    \bottomrule
  \end{tabular}
  \end{minipage}
  \hfill
  \begin{minipage}[b]{0.3\textwidth}
  \subcaption{\small EBM: \textbf{Middle (V3)}}
  \vspace{-5pt}
  \label{tab:ebm_cifar10_v3}
  \centering
  \begin{tabular}{l}
    \toprule
    Input: x, y \\
    \midrule 
    x = Conv2d(3$\times$3, 3, 32) (x) \\
    \midrule 
    x = ReLU (x) \\
    \midrule
    x = Conv2d(3$\times$3, 32, 32) (x) \\
    \midrule
    x = ReLU (x) \\
    \midrule 
    x = Maxpool(2, 2) (x) \\
    \midrule
    x = Conv2d(3$\times$3, 32, 64) (x) \\
    \midrule
    x = ReLU (x) \\
    \midrule
    x = Conv2d(3$\times$3, 64, 64) (x) \\
    \midrule
    x = ReLU (x) \\
    \midrule
    y = Embedding(N, 64) (y) \\
    \midrule 
    y = Softmax(dim=-1) (y) \\
    \midrule 
    y = y * y.shape[-1] \\
    \midrule 
    x = x * y \\
    \midrule
    x = Maxpool(2, 2) (x) \\
    \midrule
    x = FC(2304, 1024) (x) \\
    \midrule
    x = ReLU(x) \\
    \midrule
    out = FC(1024, 1) (x) \\
    \bottomrule
  \end{tabular}
  \end{minipage}

\vspace{30pt}
  \small
  \begin{minipage}[b]{0.3\textwidth}
  \subcaption{\small EBM: \textbf{End Fix Softmax (V4)}}
  \vspace{-5pt}
  \label{tab:ebm_cifar10_fix_softmax_v4}
  \centering
  \begin{tabular}{l}
    \toprule
    Input: x, y \\
    \midrule 
    x = Conv2d(3$\times$3, 3, 32) (x) \\
    \midrule 
    x = ReLU (x) \\
    \midrule
    x = Conv2d(3$\times$3, 32, 32) (x) \\
    \midrule
    x = ReLU (x) \\
    \midrule 
    x = Maxpool(2, 2) (x) \\
    \midrule
    x = Conv2d(3$\times$3, 32, 64) (x) \\
    \midrule
    x = ReLU (x) \\
    \midrule
    x = Conv2d(3$\times$3, 64, 64) (x) \\
    \midrule
    x = ReLU (x) \\
    \midrule
    x = Maxpool(2, 2) (x) \\
    \midrule
    x = FC(2304, 1024) (x) \\
    \midrule
    y = Random Projection (y) \\
    \midrule 
    y = Softmax(dim=-1) (y) \\
    \midrule 
    y = y * y.shape[-1] \\
    \midrule 
    x = x * y \\
    \midrule
    out = FC(1024, 1) (x) \\
    \bottomrule
  \end{tabular}
  \end{minipage}
  \hfill
  \begin{minipage}[b]{0.3\textwidth}
  \subcaption{\small EBM: \textbf{End Softmax (V4)}}
  \vspace{-5pt}
  \label{tab:ebm_cifar10_softmax_v4}
  \centering
  \begin{tabular}{l}
    \toprule
    Input: x, y \\
    \midrule 
    x = Conv2d(3$\times$3, 3, 32) (x) \\
    \midrule 
    x = ReLU (x) \\
    \midrule
    x = Conv2d(3$\times$3, 32, 32) (x) \\
    \midrule
    x = ReLU (x) \\
    \midrule 
    x = Maxpool(2, 2) (x) \\
    \midrule
    x = Conv2d(3$\times$3, 32, 64) (x) \\
    \midrule
    x = ReLU (x) \\
    \midrule
    x = Conv2d(3$\times$3, 64, 64) (x) \\
    \midrule
    x = ReLU (x) \\
    \midrule
    x = Maxpool(2, 2) (x) \\
    \midrule
    x = FC(2304, 1024) (x) \\
    \midrule
    y = Embedding(N, 1024) (y) \\
    \midrule 
    y = Softmax(dim=-1) (y) \\
    \midrule 
    y = y * y.shape[-1] \\
    \midrule 
    x = x * y \\
    \midrule
    out = FC(1024, 1) (x) \\
    \bottomrule
  \end{tabular}
  \end{minipage}
  \hfill
  \begin{minipage}[b]{0.3\textwidth}
  \subcaption{\small Baseline models}
  \vspace{-5pt}
  \label{tab:cls_cifar10}
  \centering
  \begin{tabular}{l}
    \toprule
    Input: x \\
    \midrule 
    x = Conv2d(3$\times$3, 3, 32) (x) \\
    \midrule 
    x = ReLU (x) \\
    \midrule
    x = Conv2d(3$\times$3, 32, 32) (x) \\
    \midrule
    x = ReLU (x) \\
    \midrule 
    x = Maxpool(2, 2) (x) \\
    \midrule
    x = Conv2d(3$\times$3, 32, 64) (x) \\
    \midrule
    x = ReLU (x) \\
    \midrule
    x = Conv2d(3$\times$3, 64, 64) (x) \\
    \midrule
    x = ReLU (x) \\
    \midrule
    x = Maxpool(2, 2) (x) \\
    \midrule
    x = FC(2304, 1024) (x) \\
    \midrule
    x = ReLU (x) \\
    \midrule 
    x = FC(1024, 1024) (x) \\
    \midrule 
    x = ReLU (x) \\
    \midrule 
    out = FC(1024, 10) (x) \\
    \bottomrule
  \end{tabular}
  \end{minipage}
\end{table*}

\begin{table*}[t]
\vspace{-8pt}
  \caption{\small{The model architectures used on the CIFAR-100 dataset. Following \citep{vandeven2020brain}, the convolutational layers were pre-trained on CIFAR-10 for all models. The `BinaryMask'-operation fully gates a randomly selected subset of $X\%$ of the nodes, with a different subset for each $y$. Hyperparameter $X$ was set using a gridsearch.}}
  \label{tab:cifar100}
  \small
  \begin{minipage}[b]{0.4\textwidth}
  \subcaption{\small The architecture of EBMs.}
  \vspace{-5pt}
  \label{tab:ebm_cifar100}
  \centering
  \begin{tabular}{l}
    \toprule
    Input: x, y \\
    \midrule 
    x = Conv2d(3$\times$3, 3, 16) (x) \\
    \midrule 
    x = BatchNorm (x) \\
    \midrule
    x = ReLU (x) \\
    \midrule
    x = Conv2d(3$\times$3, 16, 32) (x) \\
    \midrule 
    x = BatchNorm (x) \\
    \midrule
    x = ReLU (x) \\
    \midrule
    x = Conv2d(3$\times$3, 32, 64) (x) \\
    \midrule 
    x = BatchNorm (x) \\
    \midrule
    x = ReLU (x) \\
    \midrule
    x = Conv2d(3$\times$3, 64, 128) (x) \\
    \midrule 
    x = BatchNorm (x) \\
    \midrule
    x = ReLU (x) \\
    \midrule
    x = Conv2d(3$\times$3, 128, 256) (x) \\
    \midrule
    x = FC(1024, 2000) (x) \\
    \midrule
    x = ReLU (x) \\
    \midrule
    x = BinaryMask (x, y) \\
    \midrule 
    x = FC(2000, 2000) (x) \\
    \midrule 
    x = ReLU (x) \\
    \midrule
    x = BinaryMask (x, y) \\
    \midrule
    out = FC(2000, 1) (x) \\
    \bottomrule
  \end{tabular}
  \end{minipage}
  \hfill
  \begin{minipage}[b]{0.4\textwidth}
  \subcaption{\small The architecture of baseline models.}
  \vspace{-5pt}
  \label{tab:cls_cifar100}
  \centering
  \begin{tabular}{l}
    \toprule
    Input: x \\
    \midrule 
    x = Conv2d(3$\times$3, 3, 16) (x) \\
    \midrule 
    x = BatchNorm (x) \\
    \midrule
    x = ReLU (x) \\
    \midrule
    x = Conv2d(3$\times$3, 16, 32) (x) \\
    \midrule 
    x = BatchNorm (x) \\
    \midrule
    x = ReLU (x) \\
    \midrule
    x = Conv2d(3$\times$3, 32, 64) (x) \\
    \midrule 
    x = BatchNorm (x) \\
    \midrule
    x = ReLU (x) \\
    \midrule
    x = Conv2d(3$\times$3, 64, 128) (x) \\
    \midrule 
    x = BatchNorm (x) \\
    \midrule
    x = ReLU (x) \\
    \midrule
    x = Conv2d(3$\times$3, 128, 256) (x) \\
    \midrule
    x = FC(1024, 2000) (x) \\
    \midrule
    x = ReLU (x) \\
    \midrule 
    x = FC(2000, 2000) (x) \\
    \midrule 
    x = ReLU (x) \\
    \midrule 
    out = FC(2000, 100) (x) \\
    \bottomrule
  \end{tabular}
  \end{minipage}
\end{table*}

\section{More details of EBMs using replay}
\label{apx:replay_details}
In the main paper \sect{sec:classification_results}, we show that the proposed EBM formulation can be combined with existing continual learning methods, such as exact replay. In this section, we provide more experimental details of EBMs using replay.

When training a new task, we mix the new data with data sampled from a memory buffer that stores examples of previously learned tasks to train the models. 
The examples stored in the buffer are randomly selected from the classes encountered so far, and the available memory budget $k$ is equally divided over all classes encountered so far. 
On the split MNIST, permuted MNIST, and CIFAR-10 datasets, we use a memory budget of $k=1000$. On the CIFAR-100 datasets, we use a memory budget of $k=2000$.

Taking the split MNIST dataset as an example, after training the first task, we randomly select 1000 data-label pairs from the first task and save them in the replay buffer. Then we train the model on the second task. In each training batch, we randomly sample a set of data-label pairs from the second task, and randomly sample a set of data-label pairs from the replay buffer. 
We compute the final loss by adding the loss of data from the current task $\mathcal{L}_{\text{CD current}}(\boldsymbol{\theta}; \rvx, y)$ and the loss of data from the replay buffer $\mathcal{L}_{\text{CD replay}}(\boldsymbol{\theta}; \rvx, y)$ using the following equation:
\begin{equation}
    \mathcal{L}_{\text{CD}}(\boldsymbol{\theta}; \rvx, y) = 
    \mathcal{L}_{\text{CD current}}(\boldsymbol{\theta}; \rvx, y)  +
    \mathcal{L}_{\text{CD replay}}(\boldsymbol{\theta}; \rvx, y).
\label{eqn:total_loss}
\end{equation}

After we finishing training the second task, we randomly select 500 data-label pairs from the replay buffer and randomly sample 500 data-label pairs from the second task and update the replay buffer using the new sampled data-label pairs. Note we always keep the number of data-label pairs in the replay buffer at 1000. 
Then we train the model on the third task. In each training batch, we randomly sample a set of data-label pairs from the third task, and randomly sample a set of data-label pairs from the replay buffer. 
We compute the final loss using \eqn{eqn:total_loss}.
We use such a strategy to train the models until the final task. 

The baseline model, \ie ``SBC+ER'', in the main paper Table~\ref{tab:comp_baselines_boundary_aware} uses replay in the same way. The only difference is that ``SBC+ER'' uses the cross-entropy loss as described in the main paper \sect{sec:softmax} but ``EBM+ER'' uses the contrastive divergence training objective.

We use the similar training regimes for EBMs and SBC. On split MNIST, permuted MNIST, and CIFAR-10, we trained for 2000 iterations per task. On CIFAR-100, we trained for 5000 iterations per task. 
In each training batch, we sampled 128 data-label pairs from the current task and 128 data-label pairs from the replay buffer (start from the second task) to train the model.
All experiments used the Adam optimizer with learning rate $1e^{-4}$.

\section{Additional Analyses}
\label{sec:additional_analyses}

We show the model capacity comparisons in \sect{sec:text_capacity}.


\subsection{Model capacity}
\label{sec:text_capacity}
Another hypothesized reason for why EBMs suffer less from catastrophic forgetting than standard classifiers is potentially their larger effective capacity.
We thus test the model capacity of the standard classifier and EBMs on both the generated images and natural images.

\textbf{Model capacity on generated images.}
We generate a large randomized dataset of $32 \times 32$ images with each pixel value uniformly sampled from -1 to 1. Each image is then assigned a random class label between 0 and 10. We measure the model capacity by evaluating to what extent the model can fit a such dataset.
For both the standard classifier and the EBM, we evaluate three different sizes of models (small, medium, and large). 
For a fair comparison, we control the EBM and classifier have similar number of parameters.
The Small EBM and SBC have $2,348,545$ and $2,349,032$ parameters respectively. 
The medium models have $5,221,377$ (EBM) and $5,221,352$ (SBC) parameters while the large models have $33,468,417$ (EBM) and $33,465,320$ (SBC) parameters. 
The model architectures of EBMs and classifiers are shown in \tbl{tab:capacity}.

\begin{figure}[t]
\centering
\includegraphics[width=0.5\linewidth]{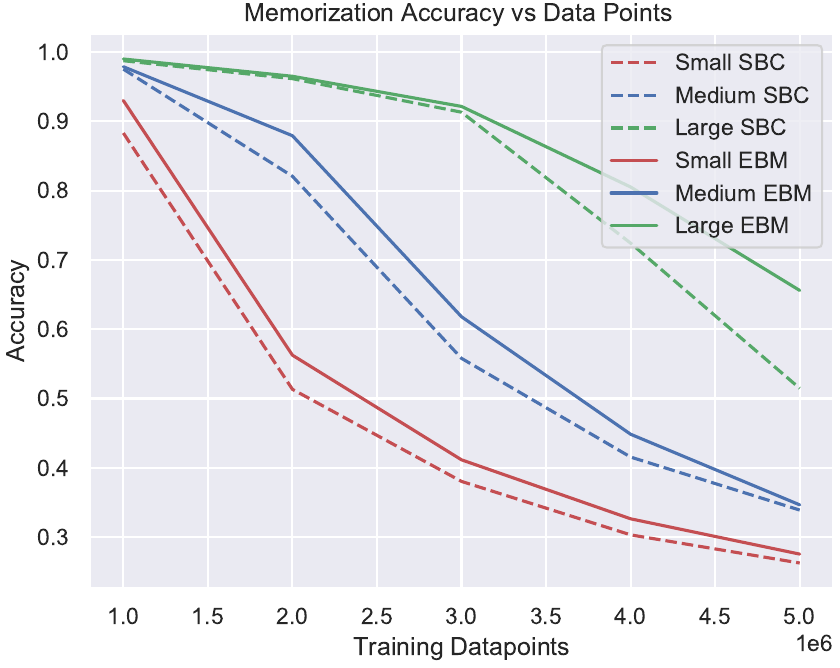}
\vspace{-5pt}
\caption{\small{Model capacity of the standard classifier (SBC) and EBM using different model sizes.}}
\label{fig:capacity}
\vspace{-5pt}
\end{figure}

The resulting training accuracies are shown in \fig{fig:capacity} with the number of data ranges from one to five millions. 
Given any number of datapoints, EBM obtains higher accuracy than the classifier, demonstrating that EBM has larger capacity to memorize data given a similar number of parameters. The gap between EBM and SBC increases when the models become larger.
The larger capacity of EBM potentially enables it to memorize more data and mitigate the forgetting problem. 

\textbf{Model capacity on natural images.}
We also compare classifiers and EBMs on natural images from CIFAR-10. Each image is assigned a random class label between 0 and 10. 
We use the same network architecture as in \tbl{tab:capacity}, but with a hidden unit size of $h=256$. 
Since there are only $50,000$ images on CIFAR-10, we use a small classifier and EBM and train them on the full dataset.
After training 100000 iterations, the EBM obtains a top-1 prediction accuracy of $82.81\%$, while the classifier is $42.19\%$. We obtain the same conclusion on natural images that EBM has a larger capacity to memorize data given a similar number of parameters.

\begin{table*}
  \caption{\small{The model architectures used for the model capacity analysis. $h$ are 512, 1024, and 4096 for the small, medium and large network, respectively.}}
  \vspace{-5pt}
  \small
  \label{tab:capacity}
  \begin{minipage}[b]{0.4\textwidth}
  \subcaption{The architecture of EBMs.}
  \vspace{-5pt}
  \label{tab:ebm_capacity}
  \centering
  \begin{tabular}{l}
    \toprule
    x = FC($32\times32\times3$, h) (x) \\
    \midrule 
    x = ReLU (x) \\
    \midrule 
    y = Embedding (y) \\
    \midrule
    x = x * y\\
    \midrule 
    x = ReLU (x) \\
    \midrule 
    out = FC(h, 1) (x) \\
    \bottomrule
  \end{tabular}
  \end{minipage}
  \hfill
  \begin{minipage}[b]{0.49\textwidth}
  \subcaption{The architecture of the standard classifier.}
  \vspace{-5pt}
  \label{tab:cls_capacity}
  \centering
  \begin{tabular}{l}
    \toprule
    x = FC($32\times32\times3$, h) (x) \\
    \midrule 
    x = ReLU (x) \\
    \midrule 
    x = FC(h, h) (x) \\
    \midrule 
    x = ReLU (x) \\
    \midrule 
    out = FC(h, 10) (x) \\
    \bottomrule
  \end{tabular}
  \end{minipage}
\end{table*}


\section{Alternative EBM Training Objective}
\label{apx:alternative_training_objective_appendix}
\subsection{Energy-based models for classification}
In the main paper \sect{sec:method_ebm_classification_cl}, we introduce an alternative EBM training objective. Here we provide more details about this training objective.
We propose to use the Boltzmann distribution to define the joint likelihood of image $\rvx$ and label $y$:
\begin{equation}
    \label{eqn:likelihood_appendix}
    \begin{split}
    p_{\boldsymbol{\theta}}(\rvx, y) 
    &= \frac{\text{exp}(- E_{\boldsymbol{\theta}}(\rvx, y))} {Z(\boldsymbol{\theta})}, 
    \\
    Z(\boldsymbol{\theta})
    &=\sum_{\rvx'\in\mathcal{X}, y'\in\mathcal{Y}} \text{exp}(- E_{\boldsymbol{\theta}}(\rvx', y'))
    \end{split}
\end{equation}
where $E_{\boldsymbol{\theta}}(\rvx, y): (\sR^D,\sN) \rightarrow \sR$ is the energy function that maps an input-label pair $(\rvx,y)$ to a scalar energy value, and $Z(\boldsymbol{\theta})$ is the partition function for normalization.

\noindent\textbf{Training.} 
We want the distribution defined by $E_{\boldsymbol{\theta}}$ to model the joint data distribution $p_D$, which we do by minimizing the negative log likelihood of the data 
\begin{equation}
\small
    \mathcal{L}_{\text{ML}}(\boldsymbol{\theta}) = 
    \E_{(x,y) \sim p_D} [ - \text{log} p_{\boldsymbol{\theta}} (\rvx, y) ].
\end{equation}
with the expanded form:
\begin{equation}
\small
    \mathcal{L}_{\text{ML}}(\boldsymbol{\theta}) 
    = 
    \E_{(\rvx, y) \sim p_D} 
    \left [
    E_{\boldsymbol{\theta}}(\rvx, y) + \log (\sum_{\rvx' \in \mathcal{X}, y' \in \mathcal{Y} } e^{-E_{\boldsymbol{\theta}}(\rvx', y')}) 
    \right].
\label{eqn:ml_loss_expend_appendix}
\end{equation}
\eqn{eqn:ml_loss_expend_appendix} minimizes the energy of $\rvx$ at the ground truth label $y$ and minimizes the overall partition function by increasing the energy of any other randomly paired $\rvx'$ and $y'$. 

\noindent \textbf{Inference.}
Given an input $\rvx$, the class label predicted by our EBMs is the class with the smallest energy at $\rvx$:
\begin{equation}
    \hat{y} = \argmin_{y' \in\mathcal{Y}} E_{\rvtheta}(\rvx, y'),
\label{eqn:ebm_inference_appendix}
\end{equation}

\subsection{Energy-based models for continual learning}
As described in the main paper \sect{sec:method_ebm_classification_cl}, directly maximizing energy across all labels of a data point $\rvx$ raises the same problem as the softmax-based classifier models that the old classes are suppressed when training a model on new classes and thus cause catastrophic forgetting.
Inspired by \citep{hinton2002training}, we find that the contrastive divergence approximation of \eqn{eqn:ml_loss_expend_appendix} can mitigate this problem and lead to a simpler equation.
We approximate \eqn{eqn:ml_loss_expend_appendix} by sampling a random pair of image $\rvx'$ and label $y'$ from the current training batch to approximate the partition function.
Our training objective is given by:
\begin{equation}
\small
    \label{eqn:cd_loss_appendix}
    \mathcal{L}_{\text{CD}}(\boldsymbol{\theta}; \rvx, y) = \E_{(\rvx,y) \sim p_D} 
    \left[ E_{\boldsymbol{\theta}}(\rvx,y) - E_{\boldsymbol{\theta}}(\rvx', y') \right],
\end{equation}
where $y$ is the ground truth label of data  $\rvx$. 

This training objective is reminiscent of the contrastive divergence training objective used to train EBMs in the main paper \eqn{eqn:cd_loss}. The major difference is that we utilize both images and labels from the current batch as our contrastive samples instead of just labels used in the main paper. 
We show in the experiments that using the proposed contrastive training objective in \eqn{eqn:cd_loss_appendix} can further improve the continual learning performance.


\subsection{Inference}
We use the same inference methods as described in the main paper to perform the \emph{Class-IL} evaluation on the continual learning datasets. The model predicts the class label $\hat{y}$ of a data point $\rvx$ from all class labels $\mathcal{Y}$, where $\rvx$ is one data point with an associated discrete label $y \in \mathcal{Y}$.
The MAP estimate is
\begin{equation}
    \hat{y} = \arg\min_{y'} E_{\theta}(\rvx, y'), \;\;\; y' \in \mathcal{Y},
\label{eqn:inference_appendix}
\end{equation}
where $E_{\theta}$ is the energy function with parameters $\theta$ after training on all the training data.

\subsection{Comparisons with existing methods}
We follow the experiments performed in the main paper and evaluate the \emph{Class-IL} on the split MNIST \citep{zenke2017continual} and permuted MNIST \citep{kirkpatrick2017overcoming} datasets on the \emph{Boundary-Aware} setting.

We compare EBMs using different training objectives and the baseline approaches in \tbl{tab:comp_baselines_boundary_aware_appendix}. 
All the baselines and EBMs use similar model architectures with similar number of model parameters for fair comparisons. 
``EBM'' means the results of the training objective used in the main paper \eqn{eqn:cd_loss}. ``EBM Alt CD'' represents the alternative training objective described in \eqn{eqn:cd_loss_appendix}.
EBMs have a significant improvement over the baseline methods on all the datasets, showing that EBMs forget less when updating models for new tasks.
``EBM Alt CD'' can further improve the continual learning performance.

\begin{table}[t]
    \caption{\small Evaluation of class-incremental learning on the \emph{boundary-aware} setting on the split MNIST and permuted datasets.
    Each experiment is performed at least 10 times with different random seeds, the results are reported as the mean $\pm$ SEM over these runs. 
    Note our comparison is restricted to methods that do not replay stored or generated data.}
    \vspace{-5pt}
    \small
    \label{tab:comp_baselines_boundary_aware_appendix}
    \begin{center}
    \setlength{\tabcolsep}{5pt}
    \scalebox{1}{
    \begin{tabular}{lcc}
    \toprule
    \bf Method & \bf splitMNIST & \bf permMNIST \\
    \midrule
    SBC & 19.90 $\pm$ 0.02 & 17.26 $\pm$ 0.19  \\ 
    \bf EBM &  \bf 53.12 $\pm$ 0.04 &  \bf 87.58 $\pm$ 0.50  \\
    \bf EBM Alt CD & \bf 60.14 $\pm$ 1.66 & \bf 89.15 $\pm$ 0.89 \\ 
    \midrule
    EWC \citep{kirkpatrick2017overcoming} & 20.01 $\pm$ 0.06 & 25.04 $\pm$ 0.50   \\
    Online EWC \citep{schwarz2018progress} & 19.96 $\pm$ 0.07 & 33.88 $\pm$ 0.49  \\
    SI \citep{zenke2017continual} & 19.99 $\pm$ 0.06 & 29.31 $\pm$ 0.62   \\
    LwF \citep{li2017learning} & 23.85 $\pm$ 0.44 & 22.64 $\pm$ 0.23  \\
    MAS \citep{aljundi2019task} &  19.50 $\pm$ 0.30 & -  \\
    BGD \citep{zeno2018task} &  19.64 $\pm$ 0.03 & 84.78 $\pm$ 1.30  \\
    \bottomrule
    \end{tabular}
    }
    \end{center}
\vspace{-5pt}
\end{table}

\clearpage



\end{document}